\documentclass[lettersize,journal]{IEEEtran}
\usepackage{amsmath,amsfonts}
\usepackage[ruled]{algorithm2e}
\usepackage{array}
\usepackage{textcomp}
\usepackage{stfloats}
\usepackage{url}
\usepackage{verbatim}
\usepackage{graphicx}
\usepackage{cite}
\usepackage{multirow}
\usepackage{xcolor}
\usepackage{caption}
\usepackage{subcaption}
\usepackage[flushleft]{threeparttable}

\newcommand{\Eref}[1]{Eq.~\ref{#1}}
\newcommand{\Tref}[1]{Table~\ref{#1}}
\newcommand{\Fref}[1]{Fig.~\ref{#1}}

\definecolor{cadmiumgreen}{rgb}{0.0, 0.42, 0.24}
\definecolor{cadmiumred}{rgb}{0.89, 0.0, 0.13}
\newcommand{\good}[1]{\tiny\textcolor{cadmiumgreen}{(+#1})}
\newcommand{\equal}[1]{\tiny(+#1)}
\newcommand{\bad}[1]{\tiny\textcolor{cadmiumred}{(\hspace{0.23ex}-\hspace{0.22ex}#1})}

\SetCommentSty{mycommfont}

\newcommand{\ours}{\textsc{Impasto}}

\DeclareMathOperator*{\argmin}{arg\,min}
\DeclareMathOperator*{\argmax}{arg\,max}

\begin{document}

\title{Imperceptible Protection against\\Style Imitation from Diffusion Models}

\author{Namhyuk Ahn, Wonhyuk Ahn, KiYoon Yoo, Daesik Kim, and Seung-Hun Nam
    \thanks{This work was partly supported by NAVER WEBTOON and by Institute of Information \& communications Technology Planning \& Evaluation (IITP) under the Leading Generative AI Human Resources Development (IITP-2026-RS-2024-00360227) grant funded by the Korea government (MSIT). This work was also supported by INHA UNIVERSITY Research Grant. \emph{(Corresponding author: Seung-Hun Nam).}}
    \thanks{Namhyuk Ahn is with the Department of Electrical and Computer Engineering, Inha University, South Korea (email: nhahn@inha.ac.kr.)} %
    \thanks{Wonhyuk Ahn, Daesik Kim, and Seung-Hun Nam are with NAVER WEBTOON AI, South Korea.}%
    \thanks{Kiyoon Yoo is with KRAFTON, South Korea.}
}


\markboth{IEEE~Transactions~on~Multimedia}
{Shell \MakeLowercase{\textit{et al.}}: A Sample Article Using IEEEtran.cls for IEEE Journals}


\maketitle

\begin{abstract}
Recent progress in diffusion models has profoundly enhanced the fidelity of image generation, but it has raised concerns about copyright infringements.
While prior methods have introduced adversarial perturbations to prevent style imitation, most are accompanied by the degradation of artworks' visual quality.
Recognizing the importance of maintaining this, we introduce a visually improved protection method while preserving its protection capability.
To this end, we devise a perceptual map to highlight areas sensitive to human eyes, guided by instance-aware refinement, which refines the protection intensity accordingly.
We also introduce a difficulty-aware protection by predicting how difficult the artwork is to protect and dynamically adjusting the intensity based on this.
Lastly, we integrate a perceptual constraints bank to further improve the imperceptibility.
Results show that our method substantially elevates the quality of the protected image without compromising on protection efficacy.
\end{abstract}

\begin{IEEEkeywords}
Image protection, style mimicry, diffusion model, imperceptible protection, adversarial perturbation
\end{IEEEkeywords}
\section{Introduction}
\IEEEPARstart{T}{he} groundbreaking advancements in large-scale diffusion models have transformed media creation workflows~\cite{rombach2022high, balaji2022ediffi, li2022upainting, podell2023sdxl, saharia2022photorealistic}.
These can be further enhanced in usability when integrated with external modules that accept multi-modal inputs~\cite{zhang2023adding,li2023gligen,mou2023t2i,huang2023composer,kim2023diffblender}.
Such innovations have also been pivotal in the realm of art creation~\cite{ko2023large,sohn2023styledrop,ahn2023dreamstyler}.
Nevertheless, generative AI, while undoubtedly beneficial, brings concerns about its potential misuse.
When someone exploits these to replicate artworks without permission, it introduces significant risks of copyright infringement.
This \textit{style imitation} becomes a serious threat to artists~\cite{shan2023glaze,vice2022artists}.
In addition, a recent survey by the Society of Authors found that 37\% of UK illustrators saw their primary‑sale income fall by 12–18\% after their styles were cloned by generative models, and 26\% had already lost work because of generative AI~\cite{soa2024survey}.
By preventing unauthorized style cloning, creators in the US and EU are more likely to retain clients who might otherwise turn to inexpensive AI-generated imitations. Meanwhile, popular IPs in Japan and Korea (e.g., manga and webtoon styles) may remain exclusive to their original artists. This also includes the protection of comic book sales, a market projected to reach approximately \$17.5 billion by 2030~\cite{comic_market}.

To counteract style imitation, previous studies have introduced adversarial perturbation~\cite{goodfellow2014explaining,madry2017towards} to artwork, transforming it into an adversarial example that can resist few-shot generation or personalization methods~\cite{van2023anti,liang2023adversarial,liang2023mist,ye2023duaw,zheng2023understanding,salman2023raising,zhao2023unlearnable}.
Specifically, building on the Stable Diffusion (SD) model~\cite{rombach2022high}, they iteratively optimize the protected (or perturbed) image to fool the diffusion models, guided by the gradients from the image encoder or denoising UNet.
While existing studies are effective in preventing style imitation, they do not prioritize the protected image's quality.
They leave discernible traces (or artifacts) on the protected images due to the inherent nature of adversarial perturbations.
Moreover, we observed that unlike adversarial attacks on classifiers~\cite{luo2022frequency,dai2023saliency}, style protection requires more intense and globally dispersed perturbations.
Consequently, despite the commendable protection performance, prior works run the risk of severely degrading the original artwork's fidelity, making it less practical for real-world applications.
While moderating the strength of protection could mitigate this, it introduces a trade-off, often compromising protection performance and making it challenging to achieve satisfactory results.

\begin{figure*}[t]
\centering
\includegraphics[width=\linewidth]{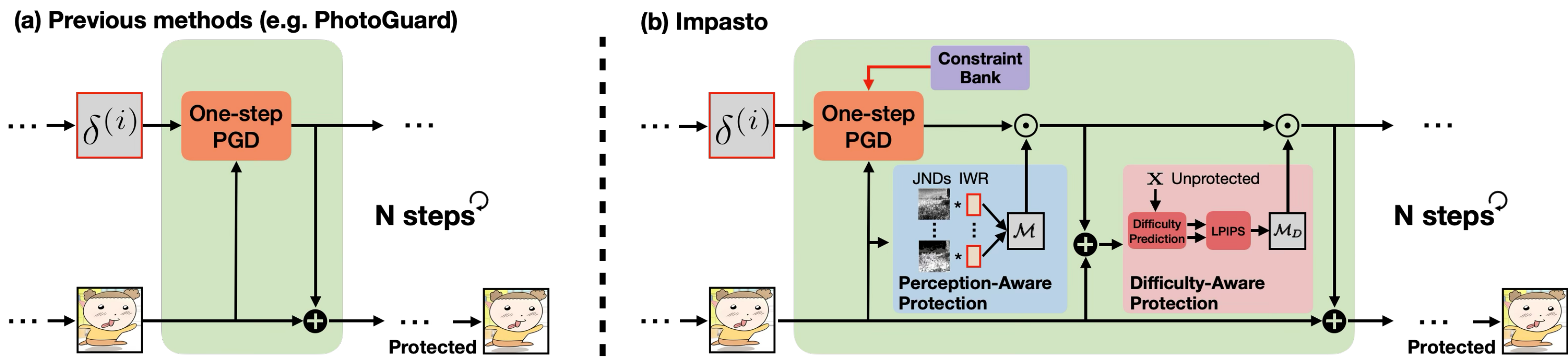}
\caption{
\textbf{Model overview.} \textbf{(a)} Current protection methods update perturbation $\delta$ with an iterative process. \textbf{(b)} On this basis, \ours\ constructs a perceptual map $\mathcal{M}$ and a difficulty map $\mathcal{M_D}$ to adaptively refine the perturbation for imperceptibility. \ours\ also employs a constraint bank in the process to achieve better imperceptibility. Red-bordered boxes indicate trainable parameters. IWR denotes instance-wise refinement.
}
\label{fig:model_overview}
\end{figure*}

To address this, we propose \ours\ (\Fref{fig:model_overview}).
We design upon the principle of \textbf{\textit{perception-aware protection}}, which focuses on perturbing regions less discernible to humans.
While many studies in the adversarial attacks regime limit perturbations to small areas to maximize imperceptibility~\cite{croce2019sparse,dai2023saliency}, they are ineffective for style protection, as personalization methods can leverage references from non-perturbed textures.
We instead adopt a \textbf{\textit{soft restriction}} strategy, which protect the entire image but with modulated intensities, relaxing the harsh condition of sparse constraints.
To implement this, it is crucial to identify which areas are perceptually more noticeable when perturbations are introduced.
For this purpose, we analyze various perceptual maps that are suitable for our task.

With a given perceptual map, the most suitable map may vary for each input image, hence a one-size-fits-all approach inherently limited in terms of imperceptibility. We instead propose a method that combines multiple perceptual maps and refines them in an image-specific manner. The \emph{instance-wise refinement} differentiates itself from previous protection methods by striking an optimal balance between imperceptibility and protection performance.
Our approach, which gradually modulates image perturbations through a soft restriction mechanism, has been explored in adversarial attack research~\cite{moosavi2016deepfool,madry2017towards}. However, we are the first to analyze restriction mechanisms in the context of image protection. Furthermore, we introduce an adaptive constraint that optimizes imperceptibility on a per-image basis, a novel contribution to the field.

For better imperceptibility, we propose a \textbf{\emph{difficulty-aware protection}}, which automatically adjusts protection strength based on how easily each region can be protected. To achieve this, we compute a difficulty map that predicts the protection difficulty of a given artwork.
Ideally, evaluating an image’s protection difficulty would require applying protection, performing a personalization method, and measuring how well the style image is mimicked. However, this process is computationally expensive, as it involves both perturbation optimization and personalization. To make this more practical, an approximation is necessary.
Our findings suggest that passing a protected image through a diffusion process~\cite{ho2020denoising,meng2021sdedit} provides a reasonable approximation. This is because diffusion process tends to amplify injected perturbations, producing patterns similar to those observed in personalization methods. By applying the diffusion process to both the original and protected images and measuring their perceptual distance, we estimate the difficulty map.
The difficulty map is then combined with the perceptual map to determine the final protection intensity.

Lastly, we introduce a \textbf{\textit{perceptual constraint bank}}, which uses various features to apply perceptual constraints. It may seem unsurprising that a constraint bank can improve imperceptibility intuitively. However, our study is the first to apply constraints across multiple spaces, improving fidelity without compromising robustness.
In addition, the perceptual constraint bank can be seen as an ensemble of multiple latent space constraints. This approach is more effective than applying constraints solely in the pixel space.

To the best of our knowledge, \ours\ is the \textbf{pioneering approach that prioritizes the protected image's quality} in the style protection task.
It achieves robust protection against style imitation and maintains visual fidelity in protected images thus it can significantly improve the trade-off balance between protection efficacy and image quality.
Moreover, \textbf{\ours\ is versatile}, enabling its attachment to any existing protection frameworks~\cite{liang2023adversarial,salman2023raising,van2023anti,liang2023mist}.
\ours\ also maintains resilience and generalizes well against a range of countermeasures and personalization techniques, performing on par with the baseline methods.
\section{Background}
\noindent\textbf{Diffusion Models.} These have gain prominence for their capacity to produce high-quality images~\cite{ho2020denoising, dhariwal2021diffusion,saharia2022photorealistic,nichol2021glide,balaji2022ediffi}.
In AI-assisted art production, Stable Diffusion (SD)~\cite{rombach2022high} is widely used because of its efficiency.
In SD, an input image $\mathbf{x}$ is projected into a latent code via an image encoder $\mathcal{E}$; $\mathbf{z} = \mathcal{E}(\mathbf{x})$.
A decoder $\mathcal{D}$ reverts the latent code to the image domain; $\mathbf{x}' = \mathcal{D}(\mathbf{z}')$.
The diffusion model derives a modified code $\mathbf{z}'$ using external factors $y$, such as text or other modalities~\cite{zhang2023adding,kim2023diffblender}.
The training objective for SD at timestep $t$ is defined as:
\begin{equation}
\mathcal{L_{SD}} = \mathbb{E}_{\mathbf{z}\sim E(\mathbf{x}), y, \epsilon\sim N(0, 1), t} [||\epsilon - \epsilon_{\theta}(\mathbf{z}_t, t, c(y))||^2_2].
\label{eq:sd}
\end{equation}
Here, a denoising UNet $\epsilon_{\theta}$ reconstructs the noised latent code $\mathbf{z}_t$, given $t$ and a conditioning vector $c(y)$.

\smallskip
\noindent\textbf{Protection Against Style Imitation.} Previous methods introduce adversarial perturbations $\delta$ to image $\mathbf{x}$, making protected image $\hat{\mathbf{x}}=\mathbf{x} + \delta$ through projected gradient descent (PGD)~\cite{madry2017towards}.
\textbf{\textit{1) Encoder-based}} methods~\cite{shan2023glaze,salman2023raising,ye2023duaw,liang2023mist} (e.g. PhotoGuard~\cite{salman2023raising} and GLAZE~\cite{shan2023glaze}) update $\delta$ under the guidance of the encoder $\mathcal{E}$.
They aim to maximize the distance between the feature of the original and protected image.
In practice, many methods minimize the distance between the protected image and target image $\mathbf{y}$ instead:
\begin{equation}
\setlength{\mathsurround}{0pt}
\setlength{\medmuskip}{0.0mu}
\setlength{\thickmuskip}{4.2mu}
\setlength{\thinmuskip}{1.2mu}
\delta = \argmin_{||\delta||_\infty\leq\eta} \, \mathcal{L_{E}}(\mathbf{x}+\delta, \;\mathbf{y}), \;\; \mathcal{L_{E}} = || \mathcal{E}(\mathbf{x} + \delta) - \mathcal{E}(\mathbf{y}) ||^2_2.
\label{eq:encoder_protet}
\end{equation}
While $L_\infty$ norm ($||\delta||_\infty\leq\eta$;\, $\eta$ is a budget) is widely used, GLAZE~\cite{shan2023glaze} adopts LPIPS~\cite{zhang2018unreasonable} and DUAW~\cite{ye2023duaw} employs SSIM~\cite{wang2004image}.
\textbf{\textit{2) Diffusion-based}} methods~\cite{van2023anti,liang2023adversarial,salman2023raising,zhao2023unlearnable,liang2023mist} (e.g. AdvDM~\cite{liang2023adversarial}) update $\delta$ under the guidance of UNet, $\epsilon_\theta$, maximizing diffusion loss $\mathcal{L_{SD}}$:
\begin{equation}
\delta = \argmax_{||\delta||_\infty\leq\eta} \, \mathcal{L_{SD}}(\mathcal{E}(\mathbf{x}+\delta)).
\label{eq:unet_protect}
\end{equation}
Upon this, Anti-DreamBooth~\cite{van2023anti} integrates DreamBooth~\cite{ruiz2023dreambooth} training in the protection optimization while Mist~\cite{liang2023mist} merges \Eref{eq:encoder_protet} and \ref{eq:unet_protect}, enhancing performance and robustness in various scenarios.
Diff-Protect~\cite{xue2023toward}, building on Mist, introduces a score distillation trick to achieve more efficient protection.
Since \ours\ is versatile and can be effortlessly integrated into any existing protection methods, we generalize the style protection as below formulation.
\begin{equation}
\delta = \argmax_{||\delta||_\infty\leq\eta} \, \mathcal{L_{SP}}(\mathbf{x}+\delta, \mathbf{y}) ,
\label{eq:sp}
\end{equation}
where $\mathcal{L_{SP}} = -\lambda_{\mathcal{E}}\mathcal{L_E}(\mathbf{x}+\delta, \mathbf{y}) + \lambda_{\mathcal{SD}}\mathcal{L_{SD}}(\mathcal{E}(\mathbf{x}+\delta))$.
Then, we employ PGD to get a protected image $\hat{\mathbf{x}}$.
Let $\mathbf{x}^{(0)}$ denote the original image.
The protected image of $i$-th optimization step is generated by a signed gradient ascent with step function sgn and step length $\alpha$ as given by:
\begin{equation}
\setlength{\thickmuskip}{2mu}
\setlength{\thinmuskip}{0mu}
\mathbf{x}^{(i)} = \Pi_{\mathcal{N}_{\eta}(\mathbf{x})} \left[ \mathbf{x}^{(i-1)} + \alpha \text{sgn} ( \nabla_{\mathbf{x}^{(i)}}\mathcal{L_{SP}}(\mathbf{x}^{(i-1)}, \; \mathbf{y}) \right],
\label{eq:pgd}
\end{equation}
where $\Pi_{\mathcal{N}_{\eta}(\mathbf{x})}$ is the projection onto $L_{\infty}$ neighborhood with radius $\eta$. 
This process is repeated $N$ steps as $\hat{\textbf{x}} = \mathbf{x}^{(N)}$.

\smallskip
\noindent\textbf{Imperceptible adversarial examples.}
The concept of imperceptibility has been actively investigated in adversarial attacks~\cite{luo2018towards,guo2018low,shen2019human,sen2019should,modas2019sparsefool,croce2019sparse,shamsabadi2020colorfool,zhao2020towards,wang2021demiguise,luo2022frequency,dai2023saliency}. Among them, some studies target specific elements; e.g. the low-frequency components~\cite{guo2018low,luo2022frequency}.
Another approach use advanced constraints~\cite{sen2019should}; color distance~\cite{zhao2020towards} or quality assessment~\cite{wang2021demiguise} are adopted.
Several methods focus on restricting perturbation regions; leveraging $L_0$ norm to produce sparse perturbation~\cite{modas2019sparsefool,croce2019sparse} or limiting perturbations to tiny salient regions~\cite{dai2023saliency}.
However, we target generative models and uniquely integrates perceptual guidance with instance-wise adaptation, which has not been explored in prior soft restriction approaches~\cite{moosavi2016deepfool,shan2023glaze}.
In addition, attacking discriminative models is fundamentally distinct from that of targeting generative models.
Hence, we employ a specialized strategy for disrupting style imitation.

\begin{figure}[t]
\centering
\newcommand{\w}{20.3mm}
\newcommand{\pad}{0.2mm}  
\centering
\raisebox{8.5mm}{\rotatebox[origin=t]{90}{\scriptsize{Example 1}}}\hspace{\pad}
\includegraphics[width=\w]{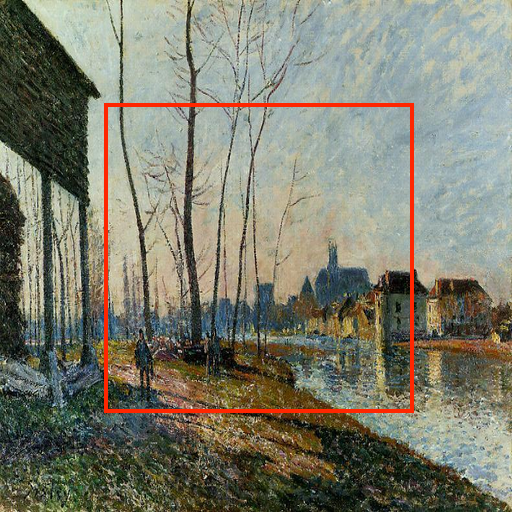}
\includegraphics[width=\w]{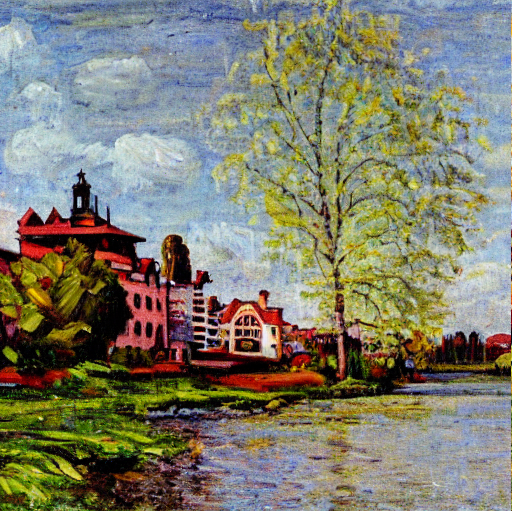}
\includegraphics[width=\w]{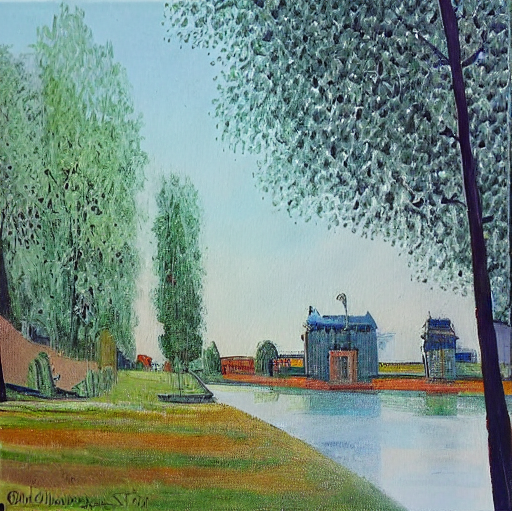}
\includegraphics[width=\w]{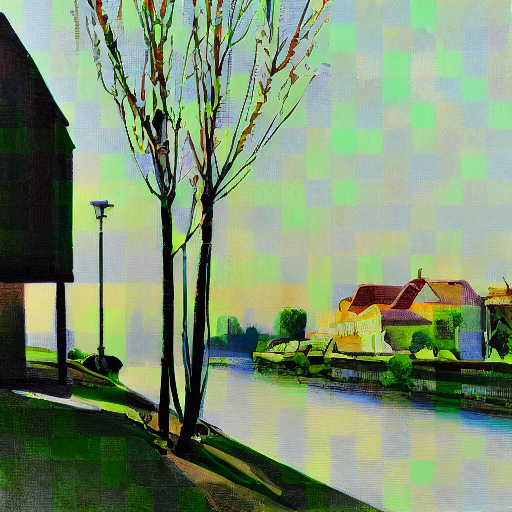}\hfill
\\
\raisebox{8.5mm}{\rotatebox[origin=t]{90}{\scriptsize{Example 2}}}\hspace{\pad} 
\includegraphics[width=\w]{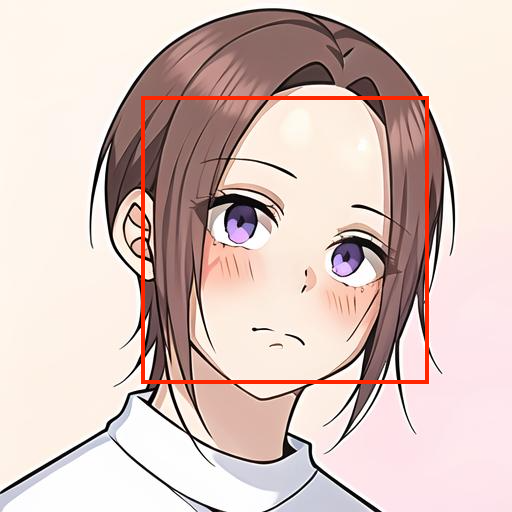}
\includegraphics[width=\w]{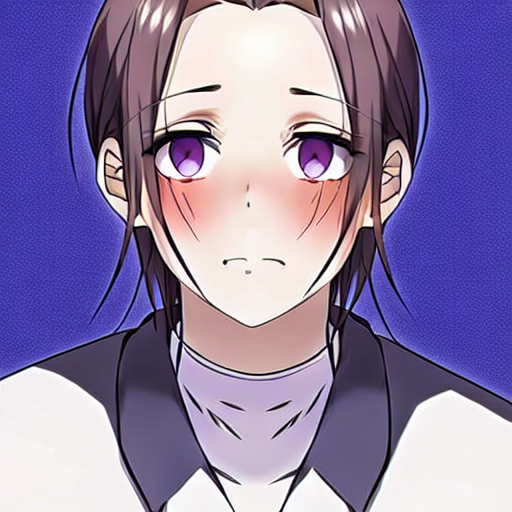}
\includegraphics[width=\w]{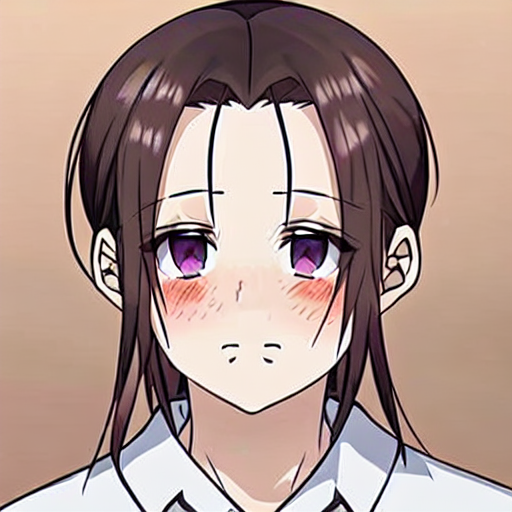}
\includegraphics[width=\w]{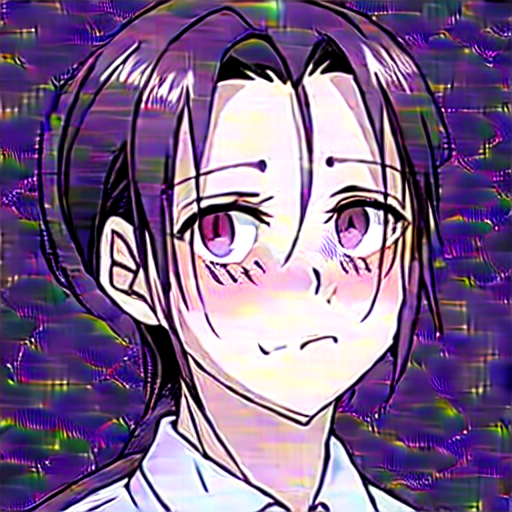}\hfill
\\
\makebox[\w][c]{\small{(a) Original}}
\makebox[\w][c]{\small{(b) No protect}}
\makebox[\w][c]{\small{(c) Partial protect}}
\makebox[\w][c]{\small{(d) Full protect}}\hfill
\caption{
\textbf{Comparison of restriction types.} (a) Original image. The red box shows the region with applied partial protection. (b–d) We compare partial and full protection by fine-tuning diffusion models on protected (or original) images with DreamBooth~\cite{ruiz2023dreambooth}. In the partial protection setting, perturbations are applied only to the central (or facial) area. The results show that partial protection is insufficient for style preservation, as personalization methods can still exploit unprotected regions to capture the artwork’s style.
}
\label{fig:center_perturbation}
\end{figure}

\begin{figure*}[t]
\begin{minipage}{0.72\linewidth}
\centering
\newcommand{\w}{15mm}
\newcommand{\h}{15mm}
\newcommand{\pad}{0.0mm}
\centering
\includegraphics[width=\w, height=\h]{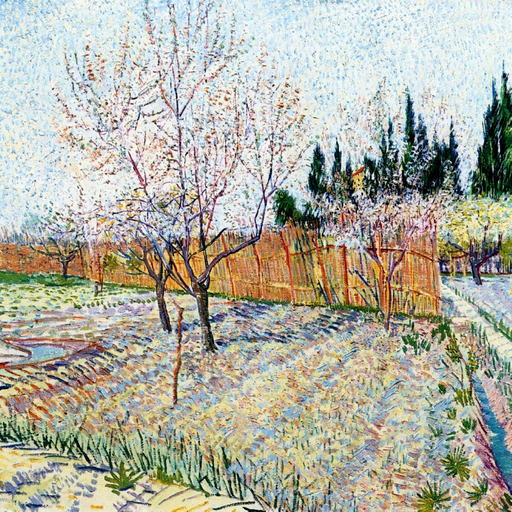}
\includegraphics[width=\w, height=\h]{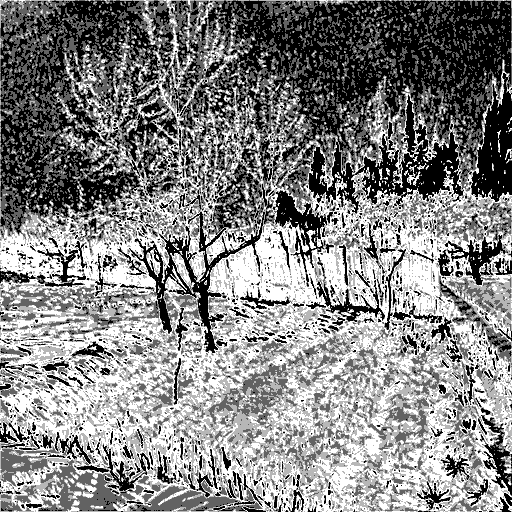}
\includegraphics[width=\w, height=\h]{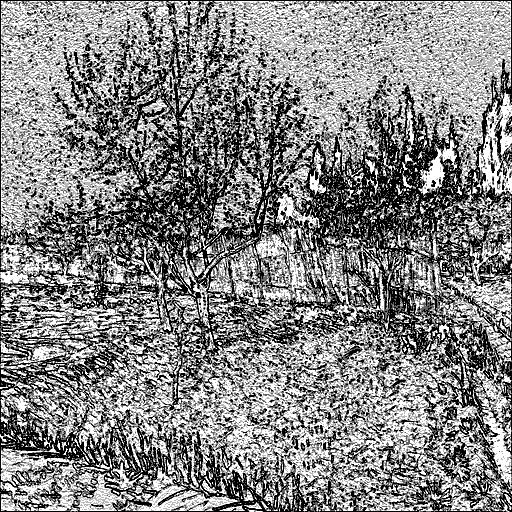}
\includegraphics[width=\w, height=\h]{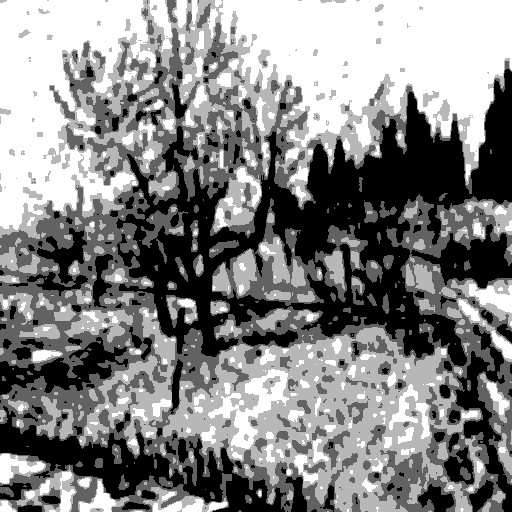}
\includegraphics[width=\w, height=\h]{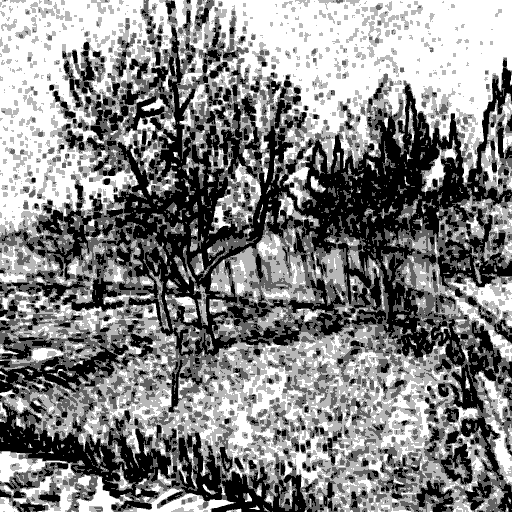}
\includegraphics[width=\w, height=\h]{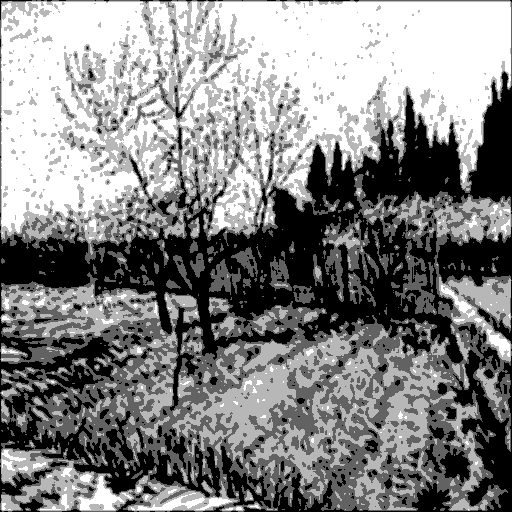}
\includegraphics[width=\w, height=\h]{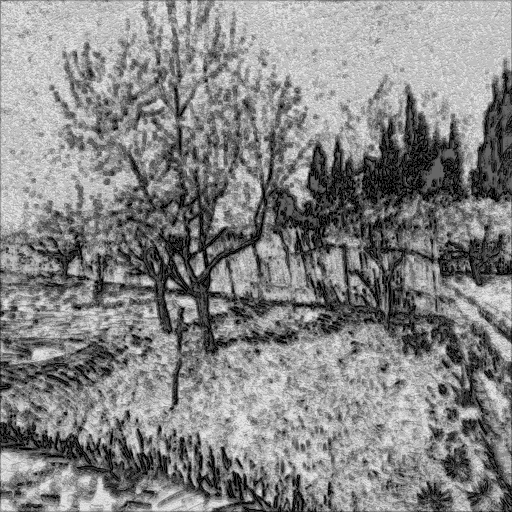}
\includegraphics[width=\w, height=\h]{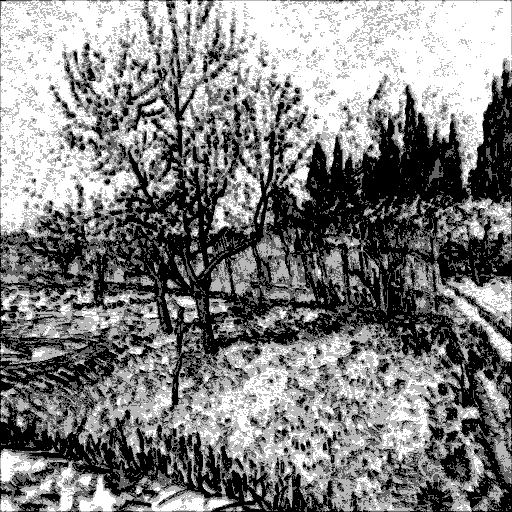}\hfill
\\
\includegraphics[width=\w, height=\h]{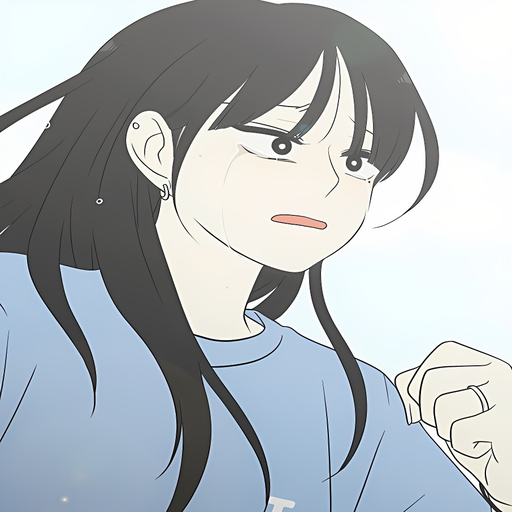}
\includegraphics[width=\w, height=\h]{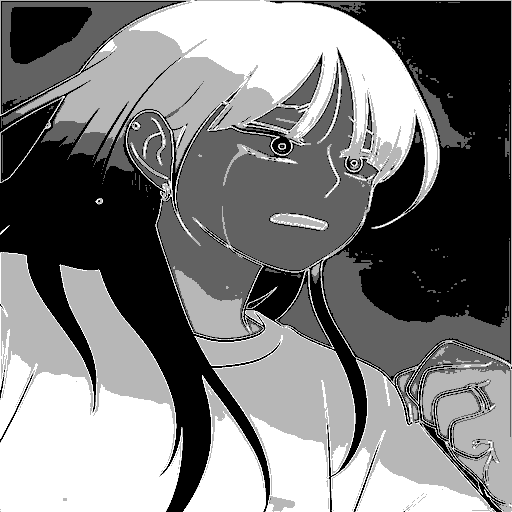}
\includegraphics[width=\w, height=\h]{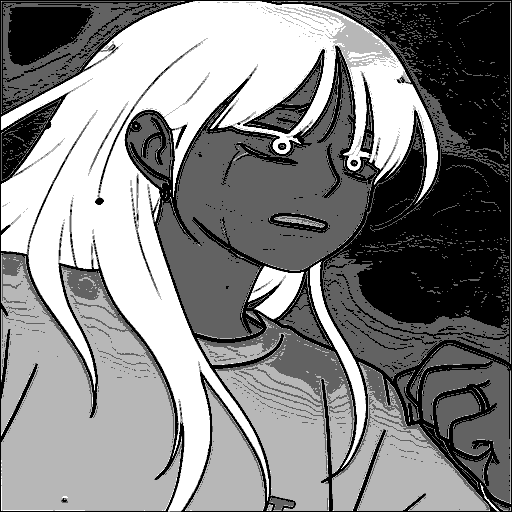}
\includegraphics[width=\w, height=\h]{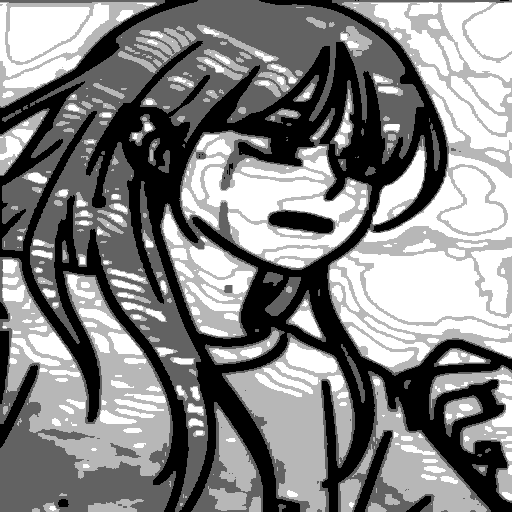}
\includegraphics[width=\w, height=\h]{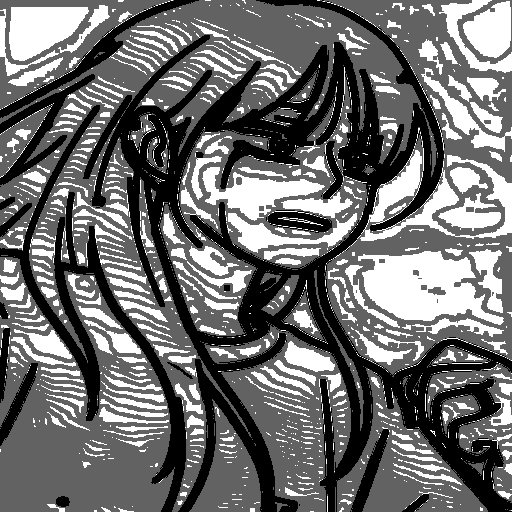}
\includegraphics[width=\w, height=\h]{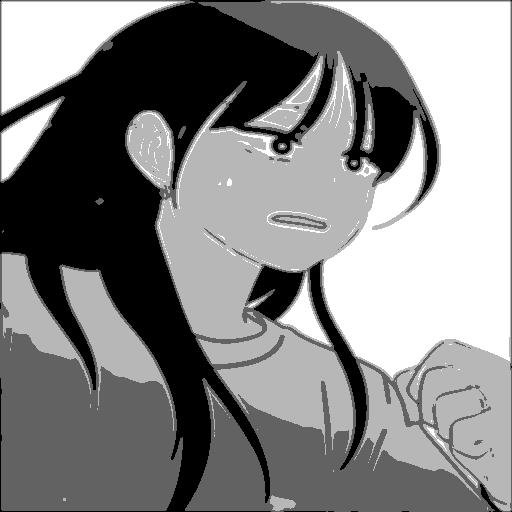}
\includegraphics[width=\w, height=\h]{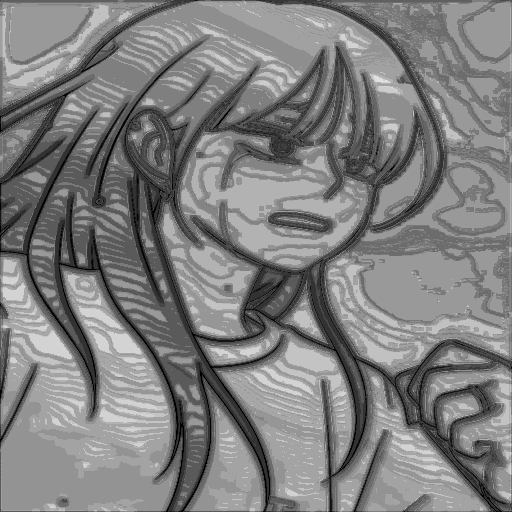}
\includegraphics[width=\w, height=\h]{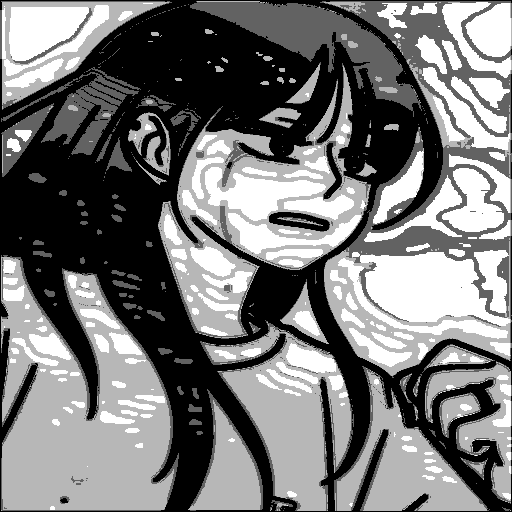}\hfill
\\
\makebox[\w][c]{\footnotesize{Input}}
\makebox[\w][c]{\footnotesize{LA}}
\makebox[\w][c]{\footnotesize{CM}}
\makebox[\w][c]{\footnotesize{CSF}}
\makebox[\w][c]{\footnotesize{Stdev}}
\makebox[\w][c]{\footnotesize{Entropy}}
\makebox[\w][c]{\footnotesize{Average}}
\makebox[\w][c]{\footnotesize{IWR}}\hfill
\end{minipage}
\begin{minipage}{0.26\linewidth}
\centering
\setlength\tabcolsep{3pt}
\footnotesize
\begingroup
\renewcommand{\arraystretch}{1.33}
\begin{tabular}{lcc}
\hline
Config. & DISTS ($\downarrow$) & FID ($\uparrow$)\\
\hline\hline
Baseline     & 0.212             & \textbf{299.1} \\
\hline
LA       & 0.175             & \underline{284.7} \\
CM       & \textbf{0.169}    & 280.8 \\
CSF      & 0.172             & 273.2 \\
Stdev     & \underline{0.171} & 277.7 \\
Entropy  & 0.174             & 277.3 \\
\hline
\end{tabular}
\endgroup
\end{minipage}
\caption{\textbf{Examples of perceptual maps.} In this study, we use luminance adaptation (LA), contrast masking (CM), contrast sensitivity function (CSF), standard deviation (Stdev), and entropy. \textbf{(Left)} Perceptual Map visualization. Darker region corresponds to increased protection intensity. \ours\ constructs perceptual map $\mathcal{M}$ by spatially averaging these estimations (Avgerage) or an learnable manner (IWR). \textbf{(Right)} Perceptual map comparison. We evaluate the protection performance of each perceptual map. Image quality is assessed through DISTS~\cite{ding2020image}, and protection performance is evaluated via FID~\cite{heusel2017gans}. The baseline is trained with Eq.~\ref{eq:pgd} and other models are trained via Eq.~\ref{eq:pap} with corresponding JNDs.}
\label{fig:jnd}
\end{figure*}
\section{Method}

\ours\ aims to enhance the imperceptibility of current image protection methods. Unlike existing approaches that rely on multi-step PGD optimization (\Fref{fig:model_overview}a), \ours\ integrates multiple imperceptibility-focused modules, allowing it to be versatile and compatible with any protection framework. As shown in \Fref{fig:model_overview}b, \ours\ consists of three core components: perception-aware protection, difficulty-aware protection, and the constraint bank.

These three modules constitute successive stages of a single pipeline that cooperatively achieve imperceptible yet robust protection: \textbf{Perception‑Aware Protection (PAP)} first decides \emph{where} to perturb by aligning noise with the human visual system, applying stronger perturbations in regions less noticeable to viewers (§\ref{sec:perception_aware_protection}). \textbf{Difficulty‑Aware Protection (DAP)} then calibrates \emph{how strongly} to perturb each image by predicting its protection difficulty and adapting the global magnitude accordingly (§\ref{sec:dap}).
Finally, the \textbf{Perceptual Constraint Bank} regularises the optimization trajectory in multiple feature spaces such as texture (with masked LPIPS), color‑frequency (with masked Low‑Pass), and semantics (with CLIP).
This provide our method the perturbations selected by PAP + DAP remain visually natural under diverse perceptual lenses (§\ref{sec:perceptual_constraint}). Together, this cascade transforms a simple PGD loop into a perception‑ and content‑adaptive procedure.

\subsection{Perception-Aware Protection}
\label{sec:perception_aware_protection}

Human observers are markedly more sensitive to noise in smooth, low-texture areas than in highly textured regions. To ensure imperceptible protection, our approach aims to identify regions where noise is least likely to be perceived and to modulate the perturbation strength accordingly on a pixel-wise basis.
We start from a straightforward yet insufficient idea of sparsely perturbing only a few pixels and gradually refine it into a soft, map-based restriction that adapts to the visual characteristics of each individual artwork. Figure~\ref{fig:center_perturbation} illustrates the progression of this design.

\smallskip
\noindent\textbf{Naive sparse restriction.}
In adversarial attacks, perturbations are often limited to sparse regions to enhance imperceptibility~\cite{croce2019sparse,dai2023saliency}. However, we found that this approach does not effectively prevent style imitation. In \Fref{fig:center_perturbation}, we compare full-image protection (Full protect) with protection that applies perturbations only to the facial region (Partial protect). The results show that partial protection fails to safeguard original artwork against DreamBooth~\cite{ruiz2023dreambooth}. We also perform a quantitative comparison of partial and full protection using FID on a painting dataset. In this test, partial protection yields an FID score of 260.2, while full protection achieves 299.1, demonstrating the superior effectiveness of full protection.
These results reveal that simply ``hiding" perturbations in a few pixels leaves large unprotected textures that diffusion-based personalization methods can still exploit.  We therefore move from an all‑or‑nothing sparse mask to a \emph{soft restriction} that redistributes perturbation strength continuously across the entire canvas, biased by perceptual visibility.

\smallskip
\noindent\textbf{Soft restriction.}
In this study, we instead adopt a soft restriction strategy.
It protects the entire image but with varying intensities across different regions.
To this end, we introduce a perceptual map, $\mathcal{M} \in \mathbb{R}^{d}$, where $d$ is the number of pixels of an image.
This map reflects the human sensitivity to subtle alterations; a value near 1.0 indicates a region with the highest perceptibility, while a value close to 0.0 signifies a region where changes are hardest to notice.
With \(\mathcal{M}\), we define perception-aware protection loss $\mathcal{L_{PAP}}(\mathbf{x}, \delta, \mathbf{y}, \mathcal{M})$, which use soft restriction-based \(L_p\) norm constraint as:
\begin{equation}
    \mathcal{L_{PAP}} =\mathcal{L_{SP}}(\mathbf{x}+\delta, \mathbf{y}) + \left(\textstyle\sum_{i=1}^{d}|\mathcal{M}_i*\delta_i|^p\right)^{1/p},
\label{eq:pap}
\end{equation}
where $*$ denotes element-wise multiplication.
Employing a soft restriction-based $L_p$ norm controls perturbations to be suppressed in regions with high perceptual visibility to humans and amplified in areas with lower visibility.
The question now is, how can we determine perceptual sensitivity?

\smallskip
\noindent\textbf{Perceptual map.}
The key to soft restriction is an accurate map of perceptual sensitivity.  We adopt just‑noticeable‑difference (JND) concept~\cite{wu2019survey} because they are explicitly designed to predict the minimum perturbation that the human eye can detect.
JND represents the minimum intensity of stimulus (perturbation in our context) required to produce a noticeable change in visual perception.
The intent behind the JND estimation model is to determine this perceptual threshold.
Given that the fundamental premise of JND--- to quantify human sensitivity to subtle changes--- aligns with our objective of perception-aware protection, we focus on analyzing the effectiveness of JNDs in our task, and which JNDs yield the good results.

To this end, as shown in \Fref{fig:jnd}, we compare five widely used JNDs: \textbf{1)} luminance adaptation (LA), which considers pixel luminance~\cite{jarsky2011synaptic}, \textbf{2)} contrast masking (CM), which measures scene complexity by luminance contrast~\cite{legge1980contrast,wu2013pattern}, \textbf{3)} contrast sensitivity function (CSF), which utilizes frequency signals~\cite{wei2009spatio}, \textbf{4)} standard deviation (Stdev), which measures spatial structure with patch-wise statistics, and \textbf{5)} entropy, which computes complexity with patch-wise entropy.
More details are described in the Supplemental Materials.

In our observation, most JND models fit surprisingly well with soft restriction (\Fref{fig:jnd}, right).
Consequently, we employ a JND-based perceptual maps, considering its simplicity and effectiveness.
Note that among JNDs we used, LA and CM emerge as superior in the quality-protection trade-off.
However, these results are the \textit{average scores} across a dataset and we note that some artworks exhibit better trade-off with other JNDs.
Especially, since artworks have diverse styles \cite{gatys2016image}, the best combinations can differ; for example, in \Fref{fig:jnd} (left), Stdev and entropy are best for top image while bottom image shows a preference for CM and CSF.
Moreover, in practical scenarios, users who protect their artwork are solely concerned with their specific pieces, not the average scores.
Thus, it is crucial to ensure that the protection method is effectively applied to \textit{every individual artwork}.

We therefore apply all the JNDs simultaneously to create the perceptual map, since relying on a single JND could result in decreased performance for some artworks.
Formally, for an artwork $\mathbf{x}$ requiring protection, we initially generate a corresponding perceptual map $\mathcal{M}$ with a collection of JNDs, $\mathbf{M} = \{M^i, \dots, M^K\}$, where $K$ is the number of JNDs.
To integrate multiple JNDs, the simplest method involves using a spatial average: $\mathcal{M} = \frac{1}{K}\textstyle\sum_{k=1}^{K} M^k$.

\smallskip
\noindent\textbf{Instance-wise refinement.} 
Although the above method is more effective in many scenarios than the single JND, applying a uniform averaged map across \emph{all artworks} may still be suboptimal for some artworks since a single, dataset‑wide JND average cannot capture the stylistic diversity of real artworks.
Moreover, the optimal perceptual map corresponding to an artwork varies depending on both artwork's structure and the applied protective perturbations.
Even for identical artwork, the detectability of perturbations is affected by changing these as human sensitivity varies with distortion type~\cite{dodge2017study}.
Considering these, we propose an \emph{instance-wise refinement (IWR)}, customizing the perceptual map $\mathcal{M}$ for each artwork.
Instance‑wise refinement re‑weights the candidate JND maps \emph{per image}, learning a convex combination that best matches that artwork’s own texture and color distribution while preserving PAP’s global imperceptibility objective.
During optimization steps, the perceptual map is refined through a weighted sum: $\mathcal{M}(\omega) = \textstyle\sum_{k=1}^{K} \text{softmax}(\mathbf{\omega})^k * M^k$, where $\mathbf{\omega}$ is refinement parameters that adjust the contributions of each JND to $\mathcal{M}$.
$\mathbf{\omega}$ is optimized using the objective function below:
\begingroup
\setlength{\jot}{0pt}
\begin{align}
\mathcal{L_M} &= || \mathcal{L_{SP}}(\mathbf{x} + \mathcal{M}' \odot \delta) - \mathcal{L_{SP}}(\mathbf{x} + \mathcal{M}(\omega) \odot \delta) ||^2_2 \nonumber\\
&+ \left(\textstyle\sum_{i=1}^{d}|\mathcal{M}(\omega)_i*\delta_i|^p\right)^{\frac{1}{p}},
\label{eq:mask}
\end{align}
\endgroup
\noindent with $\mathcal{M}'$ being the initial perceptual map before the refinement steps.
In \Eref{eq:mask}, the former term enforces the consistency of the refined perceptual map $\mathcal{M}$ with the initial map $\mathcal{M}'$ by minimizing the discrepancy in protection loss between them.
The latter term enhances the perception-aware protection for a given specific image.
As shown in \Fref{fig:jnd}, instance-wise refinement (IWR) has a pronounced effect as compared to the naive averaging, better capturing the nuances of image-specific perceptual sensitivity.

\smallskip
\noindent\textbf{Discussion.}
Restricting perturbation strength in a soft or gradual manner has been previously explored in the adversarial attack domain~\cite{moosavi2016deepfool}. However, unlike adversarial attacks, which aim to fool recognition models, our task focuses on fooling generative models, a distinction that requires further analysis—something not addressed in existing work.
In the image protection domain, similar soft restriction methods have been explored~\cite{shan2023glaze}. However, these methods did not aim to achieve more visually imperceptible protection. In addition, in our approach, we combine a perceptual map with instance-wise refinement, which, to the best of our knowledge, has not been explored in existing “soft” restriction methods. This combination is essential for robust style protection, ensuring that each artwork receives optimally distributed perturbations while maintaining imperceptibility for human viewers.

\begin{figure}[t]
\centering
\includegraphics[width=\linewidth]{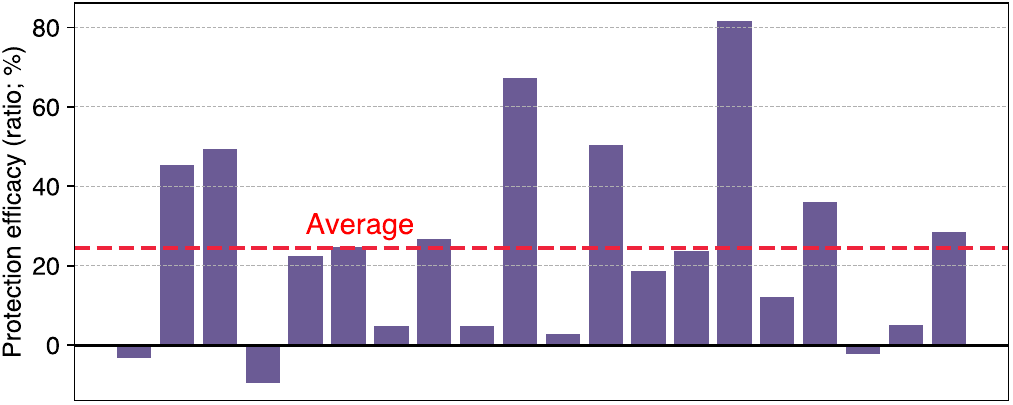}
\caption{\textbf{Protection difficulty.} We present the protection efficacy of PhotoGuard~\cite{salman2023raising} across 20 artworks (x-axis). Notably, the level of style protection varies widely between artworks, thus the average score (red dashed line) is less meaningful to users who protect their artwork.}
\label{fig:difficulty}
\end{figure}


\subsection{Difficulty-Aware Protection}
\label{sec:dap}

\noindent\textbf{Motivation.} While we adopt instance-wise refinement in PAP, it primarily focuses on ``where" to apply protection, without explicitly specifying ``how much" protection is needed. Therefore, it does not fully account for the image-specific nature.
Empirically, we observe that different artworks exhibit varying levels of tolerance to perturbations before style leakage occurs. This motivates the need for a protection method that considers both where and how strongly to apply perturbations, while maintaining imperceptibility and effectiveness.

\begin{figure}[t]
\centering
\includegraphics[width=0.9\linewidth]{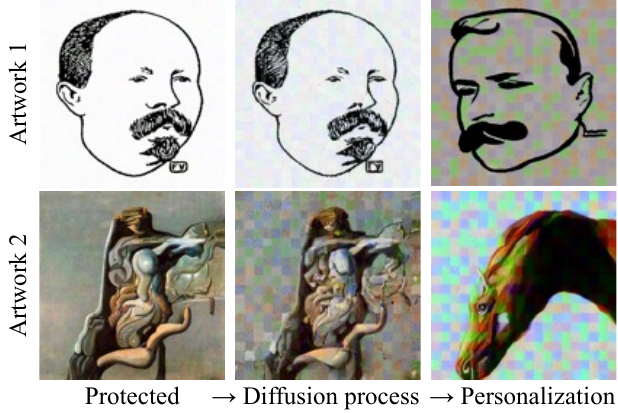}
\caption{
\textbf{Perturbation amplification.} Starting from protected images (left), the diffusion process (middle) amplifies imperceptible perturbations into visible distortions. These artifacts closely resemble those in the personalized outputs (right), suggesting that the diffusion trajectory exposes directions that personalization can exploit.}
\label{fig:dap_observation}
\end{figure}

To this end, we additionally introduce difficulty-aware protection (DAP).
Specifically, we observed that the difficulty of protection varies across images (or artworks); some images are protected properly with low protection strength, while others require higher strength to achieve the same protection level.
\Fref{fig:difficulty} shows the protection efficacy of PhotoGuard~\cite{salman2023raising} in 20 artworks. The score is calculated as $(\text{FID}_{\text{protected}}-\text{FID}_{\text{unprotected}}) / \text{FID}_{\text{unprotected}}$, with a higher value indicating better protection efficacy.
Surprisingly, the protection efficacy varies greatly across artworks. Moreover, some artworks show no difference between unprotected and protected cases. It implies that, for actual users, the \textit{average} score might be a meaningless metric. So why does the level of protection differ between artworks? This question may tie into the learning dynamics of personalization methods, which are still unclear and require further investigation.

Nevertheless, we understand that the protection performance differs between artworks, so in practice, we can adjust the protection strength accordingly when we know the \textit{difficulty}. The most straightforward way to predict the difficulty is to perform protection, personalize the protected/unprotected images with DreamBooth~\cite{ruiz2023dreambooth}, and evaluate protection efficacy as done in \Fref{fig:difficulty}. While effective, this process is extremely time-consuming, as it involves: (1) performing personalization on both clean (unprotected) and protected images, (2) comparing the results to evaluate the difficulty, and (3) reapplying protection based on the derived difficulty.

To make the above process more practical, we instead propose an approximated approach for difficulty-aware protection that utilizes the generative capacity of the same diffusion process but eliminating the need for a full personalization (\textit{i.e.} DreamBooth fine-tuning).
The core rationale stems from our preliminary empirical observations in \Fref{fig:dap_observation}: when applying forward and reverse diffusion processes to protected images, the injected perturbations tend to be amplified. Interestingly, these amplified perturbations closely resemble those exploited during DreamBooth-based personalization. In other words, if the diffusion process leads to significant perceptual changes, it implies that the corresponding regions in latent space are more vulnerable and can be easily exploited by personalization methods.
Furthermore, prior work such as \cite{xue2023toward} has also shown that injecting perturbations into input images leads to amplification in the latent space ($z$-space), thereby indicating vulnerability. Similarly, our method reveals that even a partial diffusion process effectively exposes similar vulnerabilities.
Building on this insight, we interpret such vulnerability as a proxy for protection difficulty, and accordingly adjust the protection strength based on the observed amplification during the diffusion process.

\begin{algorithm}[t]
\SetNoFillComment
\DontPrintSemicolon
\caption{Optimization of \ours}
\label{alg:ipps}
\KwData{Image $\mathbf{x}$, target image $\mathbf{y}$, perceptual maps $\mathbf{M}$}
\KwResult{Protected image $\hat{\mathbf{x}}$}
Init $\mathbf{x}^{(0)} \gets \mathbf{x}, \;\; \mathcal{M}' \gets \frac{1}{K}\textstyle\sum_{k=1}^{K} M^k, \;\; \omega \gets \frac{1}{K} \cdot \mathbf{1}^K$ \;
\For{$i=1$ \KwTo $N$}
{
    $\delta^{(i)} \gets \alpha\text{sgn} (\nabla_{\delta^{(i)}} \mathcal{L}(\mathbf{x}^{(i-1)}, \delta^{(i-1)}, \mathbf{y}, \mathcal{M})$) \;
    $\mathbf{x}^{(i)} \gets \Pi_{\mathcal{N}_\eta(\mathbf{x})} (\mathbf{x}^{(i-1)} + \delta^{(i)} \odot \mathcal{M}(\omega) \odot \mathcal{M}_D)$ \;

    \If{$i \;\text{mod}\; P == 0$}
    {
        \tcc{Instance-wise refinement}
        $\mathbf{\omega} \gets \omega - \nabla_\omega \mathcal{L_{M}}(\mathbf{x}, \delta^{(i)}, \mathbf{y}, \mathcal{M}', \mathcal{M}(\omega))$ \;
        \tcc{Difficulty-aware protection}
        $\mathcal{M}_D = \text{DAP}(\mathbf{x}, \, \mathbf{x}^{(i)})$ \;
    }
}

$\hat{\mathbf{x}} \gets \mathbf{x} + \delta^{(N)} \odot \mathcal{M}(\omega) \odot \mathcal{M}_D$ \;
\end{algorithm}

\begin{table*}[t]
\centering
\caption{
\textbf{Quantitative comparison} of protection methods w/ and w/o \ours, both selected for their comparable protection performance. \ours\ markedly elevates the protected images' quality while maintaining protection efficacy.}
\small
\setlength\tabcolsep{11pt}
\begin{tabular}{c|c|ccc|ccc}
\hline
\multirow{2}{*}{Dataset} & \multirow{2}{*}{Method} & \multicolumn{3}{c}{Protected Image Quality} & \multicolumn{3}{c}{Protection Performance} \\
\cline{3-8}
& & DISTS ($\downarrow$) & PieAPP ($\downarrow$) & TOPIQ ($\uparrow$) & NIQE ($\uparrow$) & BRISQUE ($\uparrow$) & FID ($\uparrow$) \\
\hline\hline
\multirow{8}{*}{Painting} & PhotoGuard   & 0.181 \equal{0.000} & 0.364 \equal{0.000} & 0.896 \equal{0.000} & 4.306 & 20.99 & 277.6 \\
& + \ours                                & 0.158 \good{0.023}  & 0.285 \good{0.079}  & 0.914 \good{0.018}  & 4.342 & 18.57 & 281.0 \\
\cline{2-8}
& AdvDM                            & 0.167 \equal{0.000} & 0.730 \equal{0.000} &  0.846 \equal{0.000} & 3.761 & 12.45 & 269.0 \\
& + \ours                          & 0.135 \good{0.032}  & 0.360 \good{0.370}  &  0.893 \good{0.047}  & 4.011 & 13.77 & 269.2 \\
\cline{2-8}
& Anti-DB  & 0.151 \equal{0.000} & 0.081 \equal{0.000} & 0.876 \equal{0.000} & 3.865 & 13.63 & 266.1 \\
& + \ours                                               & 0.138 \good{0.013}  & 0.008 \good{0.073}  & 0.889 \good{0.013}  & 3.780 & 12.60 & 270.9 \\
\cline{2-8}
& Mist                            & 0.167 \equal{0.000} & 0.100 \equal{0.000} &  0.846\equal{0.000} & 4.052 & 13.96 & 272.2 \\
& + \ours                                               & 0.144 \good{0.023}  & 0.077 \good{0.023}  &  0.879\good{0.033}  & 3.830 & 11.11 & 273.0 \\
\cline{2-8}
& Diff-Protect                            & 0.175 \equal{0.000} & 0.168 \equal{0.000} &  0.857\equal{0.000} & 4.146 & 15.57 & 273.8 \\
& + \ours                                               & 0.153 \good{0.022}  & 0.122 \good{0.046}  &  0.882\good{0.025}  & 4.112 & 16.20 & 275.4 \\
\hline\hline
\multirow{8}{*}{Cartoon} & PhotoGuard & 0.249 \equal{0.000} & 0.782 \equal{0.000} & 0.797  \equal{0.000} & 5.037 & 10.19 & 155.9 \\
& + \ours                                   & 0.206 \good{0.043}  & 0.706 \good{0.076}  & 0.888 \good{0.091}  & 5.620 & 11.46 & 157.7 \\
\cline{2-8}
& AdvDM                            & 0.241 \equal{0.000} & 0.776 \equal{0.000} & 0.775 \equal{0.000} & 4.802 & 10.95 & 153.5 \\
& + \ours                          & 0.231 \good{0.010}  & 0.706 \good{0.070}  & 0.888 \good{0.113} & 5.620 & 11.46 & 155.6 \\
\cline{2-8}
& Anti-DB   & 0.260 \equal{0.000} & 0.154 \equal{0.000} & 0.700\equal{0.000} & 4.437 & 15.48 & 160.6 \\
& + \ours                                               & 0.234 \good{0.026}  & 0.027 \good{0.127}  & 0.782\good{0.092}  & 4.589 & 12.35 & 161.0 \\
\cline{2-8}
& Mist                          & 0.256 \equal{0.000} & 0.160 \equal{0.000} & 0.709\equal{0.000} & 4.597 & 10.86 & 158.7 \\
& + \ours                                               & 0.238 \good{0.018}  & 0.077 \good{0.083}  & 0.772\good{0.063} & 4.693 & 11.23 & 158.3 \\
\cline{2-8}
& Diff-Protect                          & 0.258 \equal{0.000} & 0.236 \equal{0.000} & 0.758\equal{0.000} & 4.704 & 13.27 & 159.2 \\
& + \ours                                               & 0.236 \good{0.022}  & 0.145 \good{0.091}  & 0.838\good{0.080} & 4.792 & 12.28 & 160.1 \\
\hline
\end{tabular}
\label{table:comp_fix_protection}
\end{table*}

\begin{figure*}[t]
\centering
\includegraphics[width=\linewidth]{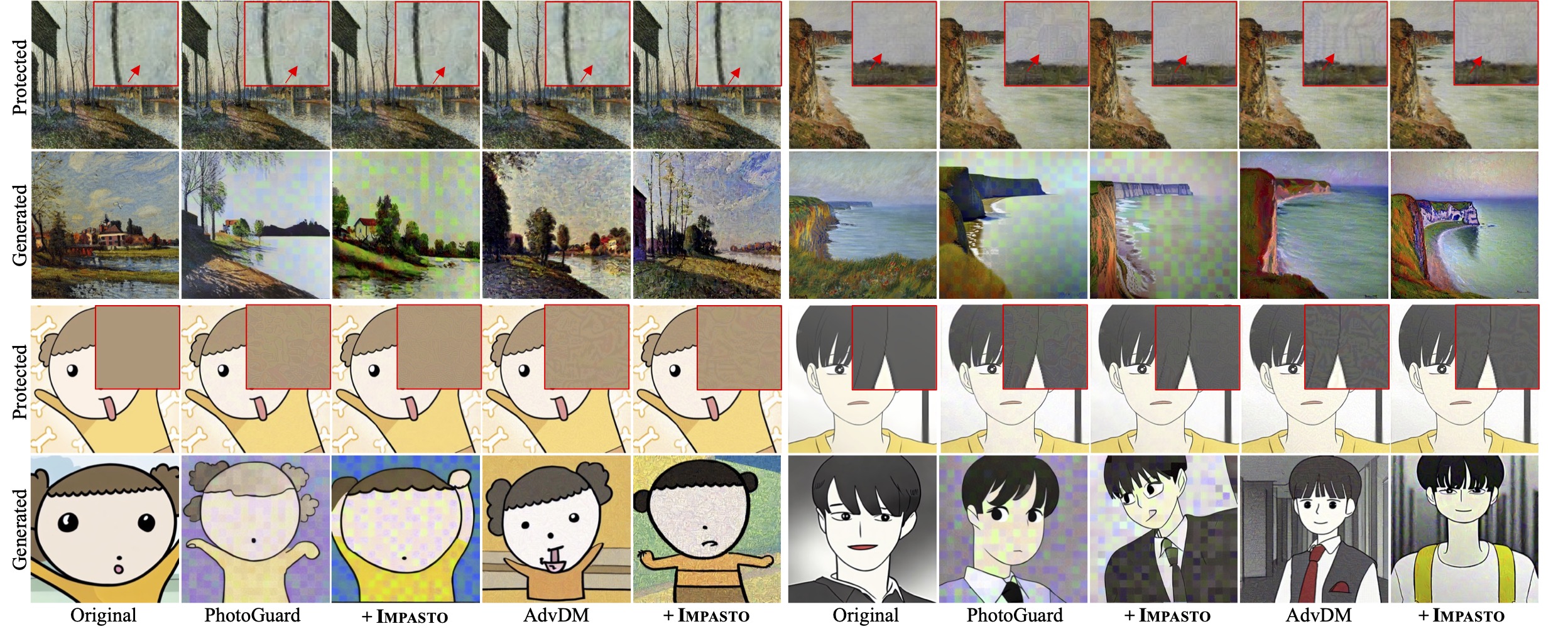}
\caption{
\textbf{Qualitative comparison.} We visualize the difference map ($\Delta$) between the protected and unprotected images.  While maintaining comparable style protection efficacy (artifacts in the generated results), \ours\ significantly enhances the quality of the protected images.
}
\label{fig:qual_comp}
\end{figure*}

\smallskip
\noindent\textbf{Approach.} 
Based on these observations, we propose an approximated approach for difficulty-aware protection (DAP) that avoids the full personalization. In essence, we estimate how easily the introduced protective perturbation can be ``undone" by a partial diffusion process~\cite{meng2021sdedit}, and use this estimate as a proxy for the protection difficulty of the image.

Specifically, given the original image $\mathbf{x}$ and the $i$-th protected image $\mathbf{x}^{(i)} = \mathbf{x} + \delta^{(i)}$, we first extract latent codes $\mathbf{z}$ and $\mathbf{z}^{(i)}$ using an encoder $\mathcal{E}$. We then apply $t$ steps of forward diffusion to each latent and perform $t$ steps of reverse denoising using a diffusion UNet $\epsilon_\theta$~\cite{meng2021sdedit}. Finally, the denoised latents are projected back to the image space using a decoder $\mathcal{D}$, yielding reconstructed images $\text{DP}(\mathbf{x})$ and $\text{DP}(\mathbf{x}^{(i)})$. This process is illustrated in \Eref{eq:dap}.
\begin{gather}
\label{eq:dap}
\text{DP}(\mathbf{x}) = \mathcal{D}(\tilde{\mathbf{z}}_0), \,\, \text{where} \\
\tilde{\mathbf{z}}_0 = \text{Reverse}(\mathbf{z}_t; \, \epsilon_\theta), \,\, \mathbf{z}_t = \text{Forward}(\mathcal{E}(\mathbf{x}), \, t). \nonumber
\end{gather}
Here, $\text{Forward}(\cdot)$ and $\text{Reverse}(\cdot)$ denote the forward and reverse diffusion processes, respectively.

Intuitively, if the reconstructed image $\text{DP}(\mathbf{x}^{(i)})$ retains a large amount of visible perturbation, it suggests that the original image $\mathbf{x}$ is relatively easy to protect. In contrast, if $\text{DP}(\mathbf{x}^{(i)})$ closely resembles $\text{DP}(\mathbf{x})$, it indicates that the protection is more difficult, as the noise was effectively reversed.
Note that we set $t = 5$ out of $T = 25$ (total denoising steps), to ensure that the diffusion process captures local perturbations without inducing major scene changes. This partial diffusion is computationally efficient and sufficiently approximates how DreamBooth may leverage or undo the added noise.

To quantify the difficulty, we compute the spatial LPIPS distance~\cite{zhang2018unreasonable} between the two outputs. LPIPS is chosen due to its strong correlation with perceptual differences in images.
\begin{equation}
\text{DAP}(\mathbf{x}, \mathbf{x}^{(i)}) = \text{LPIPS}\bigl(\text{DP}(\mathbf{x}), \, \text{DP}(\mathbf{x}^{(i)})\bigr).
\end{equation}

Consequently, $\text{DAP}(\mathbf{x}, \mathbf{x}^{(i)})$ acts as a \emph{spatial difficulty map}, which is multiplied element-wise with the perceptual map $\mathcal{M}$ to selectively reinforce protection in more vulnerable regions:
\begin{equation}
\hat{\mathbf{x}} = \mathbf{x} + \delta \,\odot\, \mathcal{M} \,\odot\, \underbrace{\text{DAP}(\mathbf{x}, \mathbf{x}^{(i)})}_{\mathcal{M}_D}.
\end{equation}

In our DAP, when $DP(\cdot)$ significantly alters the protected image compared to the original (i.e. yields a high LPIPS distance), it implies that the protection is easily neutralized by the diffusion models, thus requires stronger protection. Conversely, a low LPIPS distance suggests that the protection is resilient thus the region requires less protection strength, as personalization is less likely to succeed.
While our process does not fully replicate DreamBooth’s behavior, our empirical results show a strong correlation between the proposed surrogate measure and actual personalization performance. This enables effective, difficulty-aware protection without the computational overhead of full DreamBooth tuning.

\subsection{Perceptual Constraint Bank}
\label{sec:perceptual_constraint}
To further enhance the imperceptibility, we employ a bank of perceptual constraints across multiple feature spaces:

\smallskip
\noindent\textbf{Masked LPIPS.}
LPIPS~\cite{zhang2018unreasonable} is a widely used constraint.
Our approach distinguishes itself by applying a \textit{masked} LPIPS constraint, modulating the LPIPS influence using a perceptual map.
Let $\phi_l$ be a $l$-th layer of the LPIPS network, with a corresponding feature map resolution $d_l$, the masked LPIPS is calculated as in below equation.
\begin{equation}
\mathcal{L_{L}} = \sum_l \frac{1}{d_l} \sum^{d_l}_{i=1} \mathcal{M}_{i} * ||  w_l * (\phi_l(\mathbf{x})_i - \phi_l(\mathbf{x}+\delta)_i) ||^2_2,
\end{equation}
where $w_l$ is the channel-wise scale parameters.
By focusing on perceptually significant regions with a mask, \ours\ can achieve better protection performance.

\smallskip
\noindent\textbf{Masked low-pass.}
We also apply a pixel-domain constraint that focuses on the low-frequency components, inspired by Luo \textit{et al.}~\cite{luo2022frequency}. The loss function is as:
\begin{equation}
\mathcal{L_{LP}} = \frac{1}{d} \sum^{d}_{i=1} \mathcal{M}_{i} * || \text{LP}(\mathbf{x})_i - \text{LP}(\mathbf{x}+\delta)_i ||^2_2,
\end{equation}
where $\text{LP}(\mathbf{x})$ is the low-frequency component of image $\mathbf{x}$ (please refer to Supplemental Materials for more details).
This constraint mimics observing a painting from a distance, where perturbations in smooth regions are perceptible, while detailed textures hide these until we closely inspect the artwork.

\begin{figure}
\centering
\includegraphics[width=\linewidth]{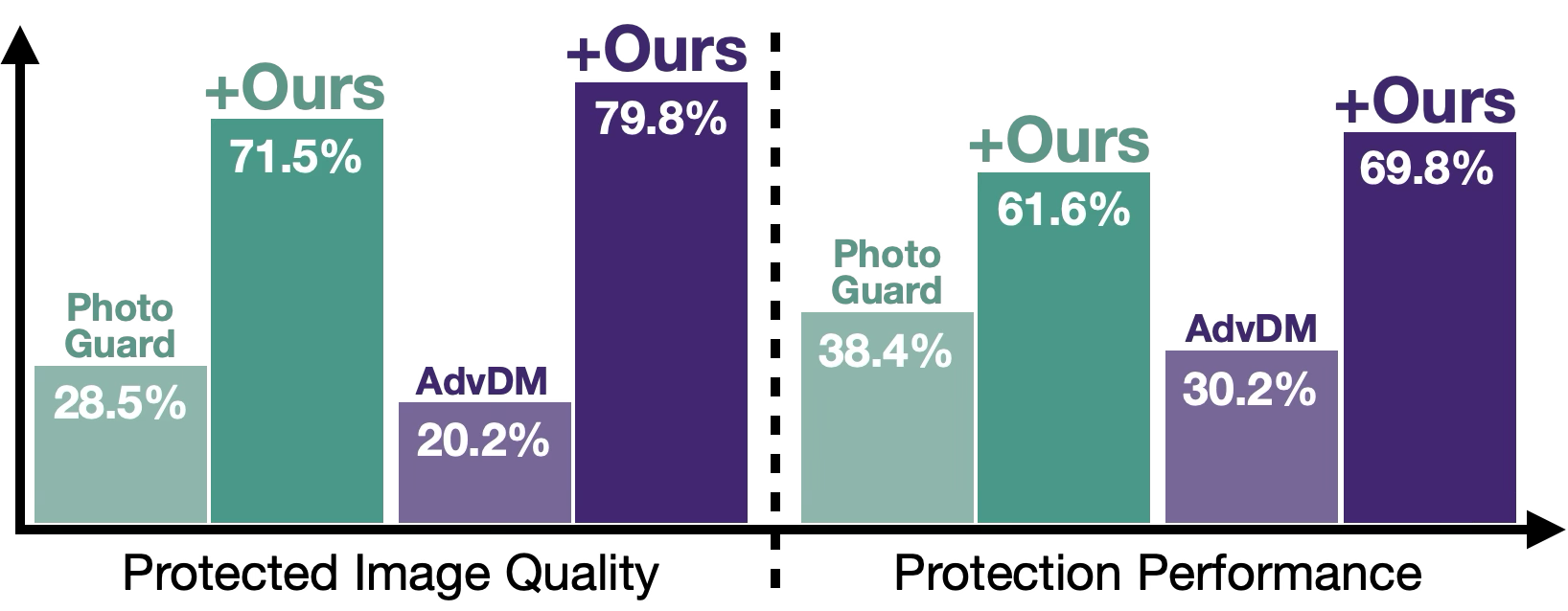}
\caption{\textbf{User preference study} (via A/B test) of PhotoGuard and AdvDM with and without \ours.}
\label{fig:user_study}
\end{figure}

\smallskip
\noindent\textbf{CLIP.}
We leverage CLIP~\cite{radford2021learning} which benefits from training on a vast and varied dataset of image-text pairs.
With the prompt $C=$ ``Noise-free image", the CLIP constraint aims to maximize the feature distance between the protected image and the descriptive prompt.
\begin{equation}
\mathcal{L_{C}} = -cos(\text{CLIP}_I(\mathbf{x}+\delta), \; \text{CLIP}_T(C)),
\label{eq:clip}
\end{equation}
where $\text{CLIP}_I, \text{CLIP}_T$ are image and text encoders.
The final protection loss, $\mathcal{L}(\mathbf{x}^{(i)}, \delta, \mathbf{y}, \mathcal{M})$ combines all the losses, weighted by their respective $\lambda$s as:
\begin{equation}
\mathcal{L} = \mathcal{L_{PAP}} + \lambda_{L}\mathcal{L_L} + \lambda_{LP}\mathcal{L_{LP}} + \lambda_{C}\mathcal{L_{C}}.
\label{eq:impaso}
\end{equation}

The perceptual constraint bank works as an ensemble of multi-feature constraints. While prior studies in adversarial attacks or image protection use constraints, they typically rely on only one or two types. Interestingly, we argue that applying constraints across multiple feature spaces can significantly enhance performance, as each feature space captures distinct aspects of the image. Despite its potential, such an approach has not been explored in previously.
Algorithm~\ref{alg:ipps} overviews the protection process of \ours.
It is designed to be versatile, allowing the integration of existing protection frameworks.

\begin{figure}[t]
\centering
\subfloat[Painting dataset]{%
    \includegraphics[width=\linewidth]{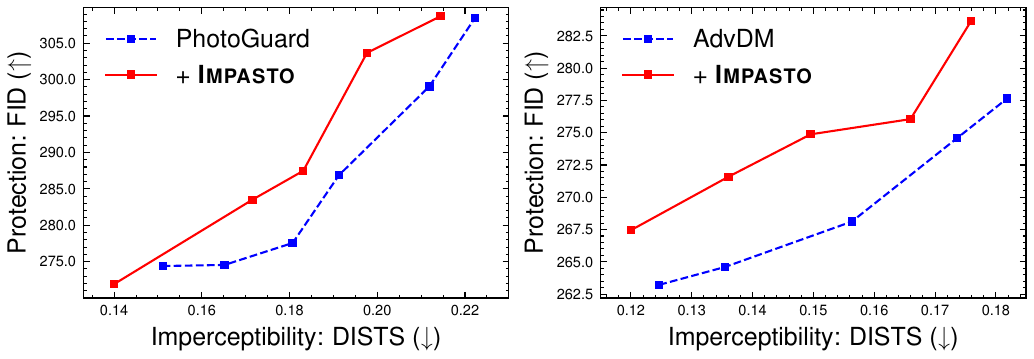}%
    \label{fig:painting_dataset}%
}\\
\subfloat[Cartoon dataset]{%
    \includegraphics[width=\linewidth]{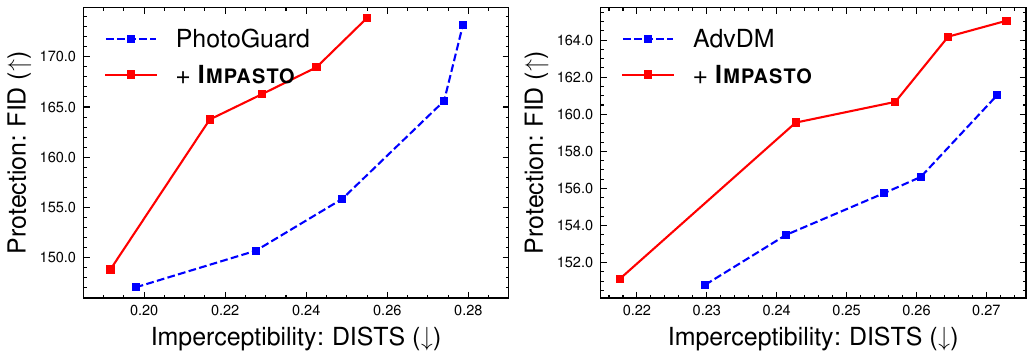}%
    \label{fig:cartoon_dataset}%
}
\caption{\textbf{Style protection comparison.} Protection performance is evaluated with FID and imperceptibility via DISTS. Adaptation of \ours\ to both PhotoGuard and AdvDM ensures superior performance.}
\label{fig:analysis_budget}
\end{figure}

\subsection{Implementation Details}
\label{sec:suppl_implementation_details}
When we implement \ours\ into the existing protection methods, we adhere to their respective settings.
Regarding the \ours's own components, we use following hyperparameters depicted in Eq.~\ref{eq:impaso} as: $\lambda_L = 5.0, \lambda_{LP} = 10.0, \lambda_C = 0.1$.
For our PAP (Eq.~\ref{eq:pap}), we utilize the $L_\infty$ norm.
On the other hand, for the IWR loss (Eq.~\ref{eq:mask}), we employ the $L_2$ norm.
When we calculate IWR loss, since the distance between $\mathcal{L_{SP}}$ respect to $\mathcal{M}'$ and $\mathcal{M}(\omega)$ (first term) is significantly smaller than the $L_2$ constraint (second term), we amplify the former term by a factor of $5\times10^7$.
For difficulty prediction in DAP, we set $t=5$ with $T=25$.
We execute IWR and DAP every $P=4$ steps.
On an A100 GPU, the overall perturbation optimization takes approximately 30 seconds per 512$\times$512 image when \ours\ is integrated with PhotoGuard. In comparison, the original PhotoGuard requires around 10 seconds under the same environment. Although our method is relatively slower, we prioritize imperceptibility; a practically critical aspect for real-world usage.

In addition, during our experiments, we observed that the magnitudes of $\mathcal{L_{E}}$ and $\mathcal{L_{SD}}$ differ considerably, with the SD loss exhibiting much smaller values.
Hence, when integrating \ours\ into UNet-based protection methods (e.g. AdvDM~\cite{liang2023adversarial}, Anti-DreamBooth~\cite{van2023anti}, Mist~\cite{liang2023mist}), we scale down $\lambda_L, \lambda_{LP}, \lambda_C$ by a factor of 0.05.
This adjustment is made to balance the influence of our proposed components, ensuring that the impact of \ours\ is consistent and effective in enhancing image protection. Last, the experiments in this paper were conducted based on SD v1.5~\cite{rombach2022high}, utilizing DreamBooth~\cite{ruiz2023dreambooth} as the method of personalization. The generalization performance of the diffusion model and personalization is provided separately in the experimental section.

\section{Experiment}

\smallskip
\noindent\textbf{Datasets.}
We utilize two art domain datasets: painting and cartoon.
The painting dataset is curated from WikiArt~\cite{tan2018improved} with a selection of 15 artists, 10 works per artist.
The cartoon dataset is a collection of 15 cartoons with 10 cartoon face images.
Further details are in Supplemental Material.

\smallskip
\noindent\textbf{Measure.}
We use DISTS~\cite{ding2020image}, PieAPP~\cite{prashnani2018pieapp} and TOPIQ~\cite{chen2023topiq} to assess protected image quality.
Protection score is measured via NIQE~\cite{mittal2012making}, BRISQUE~\cite{mittal2011blind}, and FID~\cite{heusel2017gans}.
For protection metrics, higher scores indicate stronger protection (e.g. a higher FID implies better protection performance).

It is important to note that some protection assessments, such as NIQE and BRISQUE, despite their widespread use, have limitations in our context:
1) They are designed primarily for natural image domains, making their measurements potentially unreliable in artistic domains, particularly for cartoons. 2) Traditional non-reference assessments focus on common distortions such as JPEG compression, blurring, and Gaussian noise, which may not effectively capture the unique artifacts in generated images in our task.
Therefore, these metrics tend to fluctuate instead of showing a consistent progression with varying protection strengths.
FID, although potentially inconsistent due to a limited number of evaluation samples, aligns more closely with human preferences in this task.
We suspect that it because FID measures the distance between two distributions—the original artwork and the generated image- providing more reliable assessments.
Therefore, we further validated \ours's effectiveness through human evaluation.

\begin{table}[t]
\centering
\setlength\tabcolsep{4.2pt}
\caption{\textbf{Component analysis.} PAP: perception-aware protection. w/o JND: initialize $\mathbf{M}$ with random masks, not  from JNDs. w/o Mask: Constraints without a perceptual map. DAP: difficulty-aware protection.
Base*: PhotoGuard with equal protection strength. Base**: PhotoGuard with a similar protection performance (lower protection strength).
}
\small
\begin{tabular}{lcccc}
\hline
\multirow{2}{*}{Method} & \multicolumn{2}{c}{Image Quality} & \multicolumn{2}{c}{Protection Performance} \\
\cline{2-5}
& DISTS ($\downarrow$) & TOPIQ ($\uparrow$) & NIQE ($\uparrow$) & FID ($\uparrow$) \\
\hline\hline
Base* & 0.212 & 0.845 & \underline{4.387} & \textbf{299.1} \\
\hline
+ PAP & 0.171 & 0.890                           & 4.253 & \underline{286.8} \\
w/o JND & 0.170	& 0.892 & 4.399 & 278.1 \\
\hline
+ LPIPS & 0.163 & 0.910                         & 4.369 & 280.4 \\
+ Low-pass & 0.163 & 0.911                     & 4.376 & 277.5 \\
w/o Mask & 0.163 & 0.911                      & 4.142 & 272.6 \\
+ CLIP & \underline{0.159} & \underline{0.912} & \textbf{4.479} & 279.2 \\
\hline
+ DAP & \textbf{0.157} & \textbf{0.913} & 4.342 & 281.0 \\
\hline
Base** & 0.181 & 0.896 & 4.041 & 277.6 \\
\hline
\end{tabular}
\label{table:analysis_ablation}
\end{table}

\begin{table}[t]
\centering
\small
\caption{\textbf{Component analysis.} Within \ours, we conduct an analysis by excluding the JND-based perceptual map generation, resulting in the creation of the perceptual map from random masks (denoted as w/o JND). Additionally, we evaluate the impact of omitting the perception-aware protection (denoted as w/o PAP).}
\begin{tabular}{lcccc}
\hline
\multirow{2}{*}{Method} & \multicolumn{2}{c}{Protected Image Quality} & \multicolumn{2}{c}{Protection Performance} \\
\cline{2-5}
& DISTS ($\downarrow$) & TOPIQ ($\uparrow$) & BRISQUE ($\uparrow$) & FID ($\uparrow$) \\
\hline
\hline
\ours\  & 0.159 & 0.912 & 20.74	& 279.3 \\
w/o JND & 0.158 & 0.912 & 19.30	& 268.1 \\
w/o PAP & 0.199 & 0.890 & 17.57	& 281.1 \\
\hline
\end{tabular}
\label{table:suppl_analysis_pap}
\end{table}

\smallskip
\noindent\textbf{User study.}
We conducted a user study in the form of an A/B test with a reference image as a benchmark.
A total of 60 participants were involved, tasked with determining the better-quality protected image (given the reference original artwork) and identifying the lower-quality generated image (given a reference DreamBooth~\cite{ruiz2023dreambooth} generated image).
Each participant was asked to vote on 8 questions for PhotoGuard~\cite{salman2023raising} and another 9 for AdvDM~\cite{liang2023adversarial}.

\smallskip
\noindent\textbf{Baseline.}
\ours\ can be incorporated into any existing protection method. Therefore, we apply \ours\ to both encoder and diffusion-based protection frameworks. In our model comparison, we evaluate the performance of methods with and without \ours, using PhotoGuard~\cite{salman2023raising}, AdvDM~\cite{liang2023adversarial}, Mist~\cite{liang2023mist}, Diff-Protect~\cite{xue2023toward}, and Anti-DreamBooth~\cite{van2023anti} as baselines.

\subsection{Model Comparison}
Prior work compares protection methods under a fixed perturbation strength. However, we observe that both protection efficacy and imperceptibility can vary considerably depending on the specific loss function employed. For example, PhotoGuard~\cite{salman2023raising} and AdvDM~\cite{liang2023adversarial} follow distinct optimization objectives, leading to different trade-off patterns between visibility and robustness. Furthermore, focusing solely on protection performance without considering visual imperceptibility provides an incomplete picture that may not align with real-world deployment needs. To account for this, we adaptively adjust the perturbation strength for each method \emph{to achieve comparable protection levels} (similar protection efficacy).

\Tref{table:comp_fix_protection} presents a comparison of image quality under a comparable protection performance of the models with and without \ours.
Notably, \ours\ achieves comparable or even superior protection strength while significantly improving visual quality, thereby offering a more favorable trade-off between robustness and invisibility.
Across all the scenarios, \ours\ substantially enhances the fidelity of the protected images.
\Fref{fig:qual_comp} also supports the superior efficacy of \ours;
it successfully minimizes artifacts, in contrast to baselines that leave discernible traces on the artwork.
It is particularly pronounced in the cartoon dataset (bottom), where both vanilla methods introduce noticeable artifacts in facial areas, potentially disrupting user immersion when reading cartoons.
In contrast, \ours\ reduces artifacts significantly, rendering them nearly invisible unless examined closely and meticulously.
This suggests that our method ensures stronger protection with minimal perceptual degradation, which is especially beneficial for sensitive visual domains such as character-centric content.
User evaluation further confirms the effectiveness of \ours\ (\Fref{fig:user_study}).

\begin{table}[t]
\centering
\setlength\tabcolsep{4.9pt}
\small
\caption{\textbf{Perception-aware protection.} The efficacy of JNDs (LA and CM as they show best results) is presented along with an averaged perceptual map and with IWR. IWR$^\dagger$: IWR is conducted before optimization.}
\begin{tabular}{lcccc}
\hline
\multirow{2}{*}{Method} & \multicolumn{2}{c}{Image Quality} & \multicolumn{2}{c}{Protection Performance} \\
\cline{2-5}
& DISTS ($\downarrow$) & TOPIQ ($\uparrow$) & NIQE ($\uparrow$) & FID ($\uparrow$) \\
\hline\hline
Baseline & 0.212 & 0.845 & \textbf{4.387} & \textbf{299.1} \\
\hline
LA       & 0.175 & 0.879          & 4.298 & 284.7 \\
CM       & \textbf{0.169} & 0.882          & 3.986 & 280.8 \\
\hline
Average      & \underline{0.170} & \underline{0.891} & 3.971 & 277.9 \\
IWR$^\dagger$ & 0.171 & \textbf{0.894} & 4.016 & 282.5 \\
IWR & 0.171 & 0.890 & \underline{4.306} & \underline{286.8} \\
\hline
\end{tabular}
\label{table:analysis_map}
\end{table}

\smallskip

\noindent\textbf{Varying strengths.}
As demonstrated in \Fref{fig:analysis_budget}, we manipulate the protection strengths (budget) to delineate an imperceptibility-protection trade-off curve.
On both painting and cartoon datasets, employing \ours\ into the baselines considerably improves the trade-off dynamics.
Specifically, \ours\ consistently yields higher image quality at the same protection level, or stronger protection at the same perceptual budget, highlighting a more favorable imperceptibility-robustness balance compared to existing methods.

\subsection{Model Analysis}

\noindent\textbf{Ablation study.}
As shown in \Tref{table:analysis_ablation}, the perception-aware protection (PAP) markedly improves image fidelity over Photoguard (Base*), albeit with a slight reduction in protection efficacy.
However, compared to Photoguard with lower strength (Base**), PAP achieves better preservation of protection performance with enhanced image quality.
We observed that when $\mathcal{M}$ is formed from a random mask, not JND, (w/o JND), the protection performance is degraded.
Employing a constraint bank that combines LPIPS, low-pass, and CLIP constraints significantly improves image quality while maintaining protection performance comparable to Photoguard (Base**). LPIPS emphasizes perceptual similarity, the low-pass filter suppresses high-frequency noise, and CLIP ensures semantic consistency. These components work synergistically to balance visual refinement and robust protection, validating the method’s effectiveness.
The omission of the perceptual map $\mathcal{M}$ in the constraints (w/o Mask) leads to a decline in protection performance, akin to the observations in the PAP case.
The difficulty-aware protection (DAP) also 
enhances both image quality and protection efficacy (FID).

One might expect that PAP with JND maintains protection scores while improving image quality.
Indeed, the random mask (PAP without JND) constrains the protection intensity in a similar level to the JND-based map.
This is because both maps are normalized between 0 and 1.
As a result, they yield similar perturbation magnitudes, leading to a comparable image quality.
Nonetheless, the JND-based perceptual map prioritizes less sensitive regions for perturbation while reducing it in highly sensitive areas.
Despite the similar protection magnitudes, the JND-based PAP achieves more effective protection. 
This hints at the possibility that the JND maps are helpful in finding the areas that are important for style protection.
For instance, in areas with complex textures, the PAP applies more perturbations than its non-JND counterpart, yet these perturbations remain imperceptible to humans, thereby ensuring perceptually acceptable image quality. Moreover, the interaction between the JND-based perceptual map and PAP drives a synergistic effect, where the former guides the spatial distribution of perturbations and the latter enforces robust protection, collectively enhancing the overall protection efficacy.

In \Tref{table:suppl_analysis_pap}, we show additional analysis by omitting certain components from the full \ours\ framework.
Specifically, we assess the impact of 1) removing the JND-based perception map, which results in initializing $\mathbf{M}$ with random masks (w/o JND), and 2) excluding the proposed perception-aware protection (PAP), thereby relying solely on the perceptual constraint bank (w/o PAP).
For the w/o JND case, in contrast to \Tref{table:analysis_ablation}, we start with a full-component model and remove only the JND part.
Results indicate that discarding the JNDs leads to a decreased protection performance with a similar image quality, corroborating the findings in \Tref{table:analysis_ablation}.
When PAP is not applied, there is a slight increase in protection performance (FID: 279.3 $\rightarrow$ 281.1) with significant compromise on image quality.
These observations highlight that the combined interaction of PAP and the JND-based perceptual map is instrumental in achieving an imperceptible style protection framework by efficiently balancing high protection performance with superior image quality.

\begin{figure}[t]
\centering
\begin{subfigure}{.24\textwidth}
    \includegraphics[width=\linewidth]{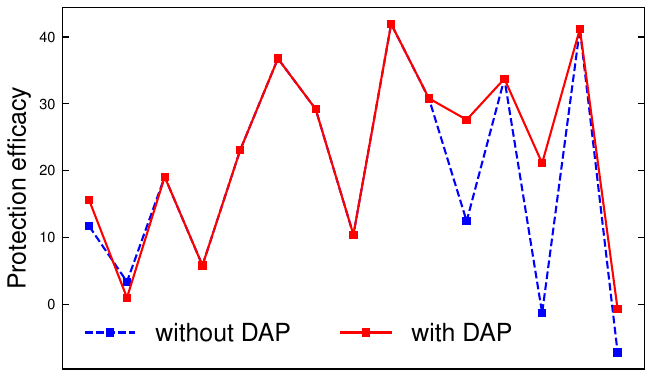}
    \caption{PhotoGuard + \ours}
\end{subfigure}
\begin{subfigure}{.24\textwidth}
    \includegraphics[width=\linewidth]{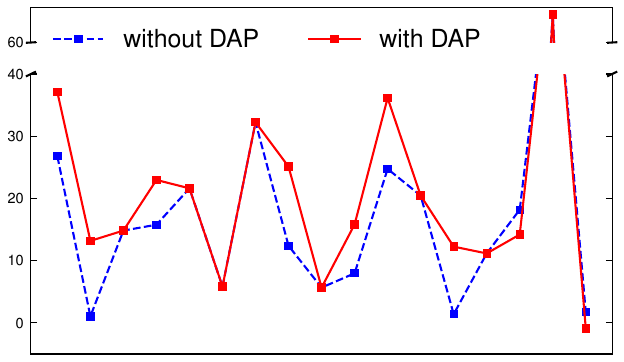}
    \caption{AdvDM + \ours}
\end{subfigure}
\caption{\textbf{Difficulty-aware protection.} We compare on the Cartoon dataset with and without the difficulty-aware protection (DAP). Higher values indicates better protection.}
\label{fig:difficulty_result}
\end{figure}

\smallskip
\noindent\textbf{JNDs.}
Here, we further discuss the protection performance of JND-based perception map depicted in \Fref{fig:jnd} (right).
When compared to the baseline, perceptual map improves image quality (DISTS) across the board, but it also leads to a compromise in protection performance (FID).
Among the JNDs, LA demonstrates the best protection performance.
We conjecture that in many artworks, the majority of areas fall high or low-luminance, thereby maintaining perturbations strength high across extensive regions.
However, as shown in upper image, even simple textures (\textit{e.g.} sky) can have strong perturbations, placing LA at the lower image quality.
On the other hand, CSF, Std, and Entropy generally apply high perturbations only to specific areas, such as edges, resulting in most regions being not fully protected and consequently, causing a huge degradation in protection performance.
CM's protection intensity is also determined by spatial changes but this covers more detailed local regions (see \textit{fields} in upper image) and also being based on contrast, leading to both high quality and effective protection.

\smallskip
\noindent\textbf{Perceptual map.}
We analyze the perception-aware protection in \Tref{table:analysis_map}.
LA and CM enhance the protected image's fidelity but at the cost of compromising protection performance.
Multiple JNDs through averaging leads to better image quality, but it also reduces the protection performance.
We speculate that such a straightforward averaging method may not adequately capture the unique structural elements of the image, resulting in overly smooth perturbations that could weaken the protection performance.
On the other hand, IWR enhances all protection scores while preserving satisfactory image quality, as it can adapt to the specific textures and structures of a given artwork.
It's noteworthy that applying IWR prior to perturbation optimization (IWR$^\dagger$), where it does not consider the perturbations, slightly diminishes protection performance, accentuating the importance of modeling the interplay between artwork and applied perturbation to finalization mask $\mathcal{M}$.

\begin{figure}[t]
\centering
\includegraphics[width=\linewidth]{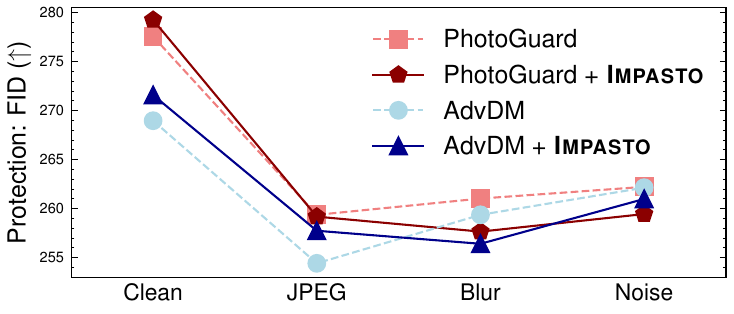}
\caption{\textbf{Robustness evaluation.} Methods with \ours\ exhibit comparable protection performance to baselines.}
\label{fig:countermeasure}
\end{figure}

\begin{table}[t]
\centering
\small
\caption{\textbf{Evaluating the impact of varying refinement interval step $P$.} In this analysis, we assess the impact on performance by varying the refinement interval step $P$ for executing IWR and DAP.}
\begin{tabular}{lcccc}
\hline
\multirow{2}{*}{Method} & \multicolumn{2}{c}{Protected Image Quality} & \multicolumn{2}{c}{Protection Performance} \\
\cline{2-5}
& DISTS ($\downarrow$) & TOPIQ ($\uparrow$) & BRISQUE ($\uparrow$) & FID ($\uparrow$) \\
\hline
\hline
$P=1$  & 0.159 & 0.913 & 19.29	& 278.4 \\
$P=2$  & 0.158 & 0.913 & 18.19	& 279.7 \\
$P=4$ & 0.159 & 0.912 & 20.74	& 279.3 \\
$P=10$  & 0.159 & 0.913 & 18.49	& 274.3 \\
\hline
\end{tabular}
\label{table:suppl_analysis_iwr}
\end{table}

\begin{table}[t]
\centering
\small
\setlength\tabcolsep{15pt}
\caption{\textbf{Comparison on the blur countermeasure} of PhotoGuard with and without \ours\ at a similar level of protected image quality.}
\begin{tabular}{ccc}
\hline
FID ($\uparrow$) & PhotoGuard & + \ours \\
\hline\hline
Clean & 274.4 & \textbf{279.3} \\
Blur & 273.5 & \textbf{274.6} \\
\hline
\end{tabular}
\label{tab:suppl_countermeasure}
\end{table}

\smallskip
\noindent\textbf{Difficulty-aware protection (DAP).}
In \Fref{fig:difficulty_result}, we compare the protection performance with and without DAP evaluated on the Cartoon dataset. \Fref{fig:difficulty_result}a and b show \ours\ combined with PhotoGuard and AdvDM, respectively.
Compared to cases without DAP, the use of DAP results in similar or improved protection efficacy for most test samples. Notably, for cases with higher protection difficulty, DAP naturally selects a stronger budget, leading to enhanced protection efficacy.

However, as DAP approximates the actual difficulty, failure cases may arise. Measuring actual protection difficulty requires: (1) performing personalization on unprotected images to obtain results, (2) protecting images with Impasto (w/o DAP), and (3) comparing the results of personalization on protected and unprotected images. This process also involves an additional perturbation optimization step based on the measured difficulty, which is impractical. Instead, DAP approximates difficulty, which can result in inaccuracies. For example, the second sample in \Fref{fig:difficulty_result}a and the third and first samples from the right in \Fref{fig:difficulty_result}b illustrate such cases.
To evaluate the gap between actual difficulty and our approximation, we test PhotoGuard + \ours\ on the Cartoon dataset using the actual difficulty measurement process. The FID score for our DAP (approximation) is 157.7, while the actual difficulty measurement yields 162.5. While the approximated approach is not perfect, it remains effective for practical purposes, given the challenges of measuring real difficulty.

\smallskip
\noindent\textbf{Varying refinement interval step.}
In this analysis, we explore the effect of varying the refinement interval step $P$ on instance-wise refinement (IWR) and difficulty-aware protection (DAP) to evaluate changes in protection performance (\Tref{table:suppl_analysis_iwr}). We observe that even as $P$ increases from 1 to 4 (i.e., with less frequent refinements), both image quality and protection capability remain relatively stable. This stability can be attributed to the initial use of a JND-based perceptual map $\mathcal{M}$, which requires only minor adjustments during the IWR step. However, when $P$ is increased to 10, a significant reduction in protection performance is observed, indicating that such a large interval is insufficient for optimal refinement. Therefore, we set $P=4$ as the default, balancing performance efficacy with the number of optimization steps.

\smallskip
\noindent\textbf{Countermeasures.}
To analyze the robustness of \ours, we conduct countermeasure experiments with JPEG compression ($q=40$), Gaussian blur ($3\times3$ kernel, $\sigma=0.02$), and Gaussian noise ($\sigma=0.02$).
Results indicate a performance degradation of all protection methods when these countermeasures are applied, as they tend to remove the protective perturbations (\Fref{fig:countermeasure}).
Nonetheless, \ours\ demonstrates comparable robustness against such countermeasures.

For the blur countermeasure, there is a slight performance degradation with \ours, as subtle perturbations are particularly weakened to this. However, we observed that baselines (both Photoguard and AdvDM) with low-budget protection are also vulnerable to blur.
The reason for this phenomenon is that the blurring effect removes most high-frequencies, making weak perturbations less effective. We note that if baseline protection models are applied with weak protection strength (to match the protected images’ perceptual quality to ours), they also exhibit degraded protection performance.

\begin{figure}
\centering
\includegraphics[width=\linewidth]{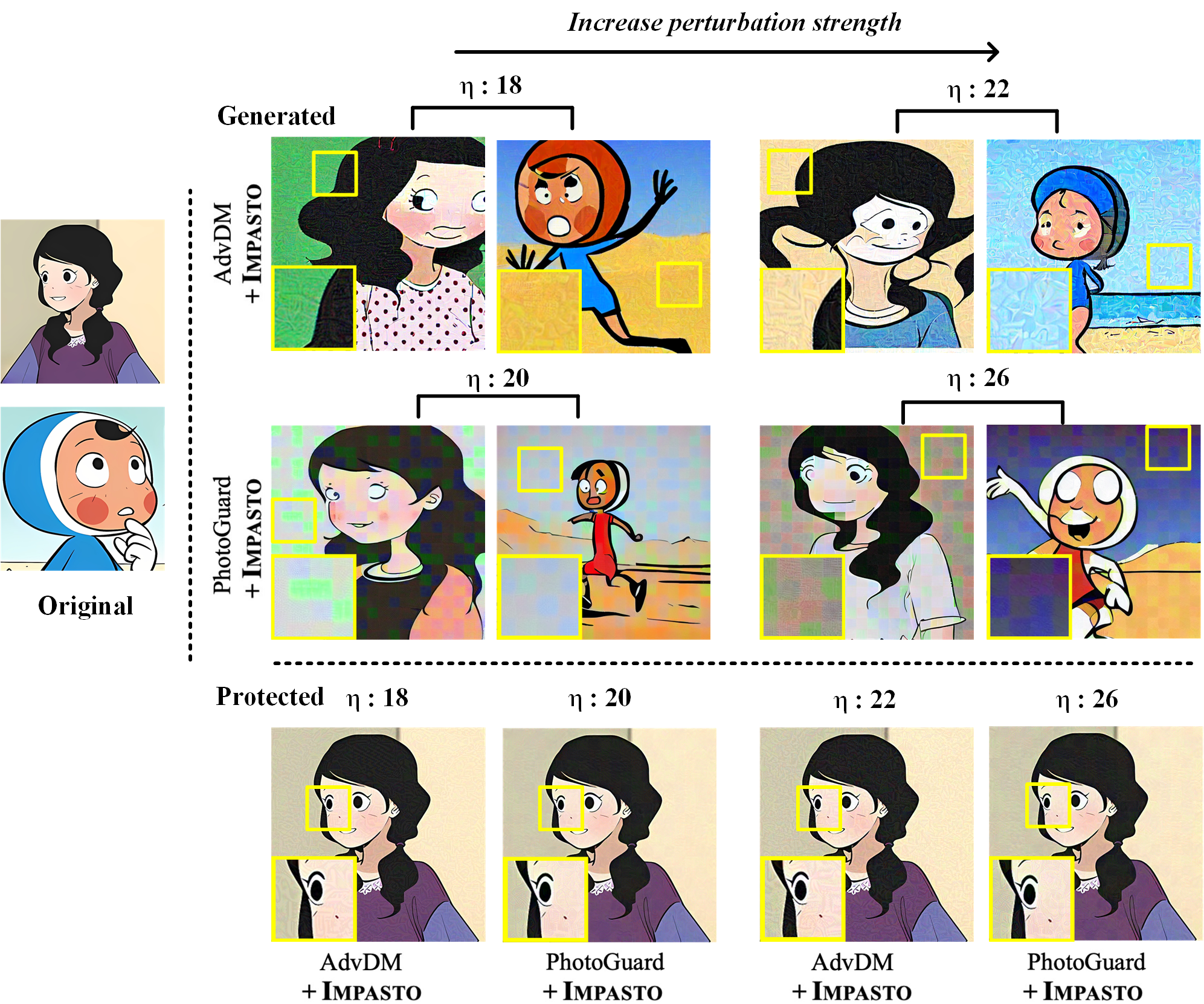}
\caption{
\textbf{Degradation tendency} of PhotoGuard and AdvDM compared to \ours~as perturbation strength increases. (Top-left) Reference original image. (Top-right) Images generated after fine-tuning diffusion models on inputs with increasing levels of protection strength (more perturbations). (Bottom-right) Protected images with increasing levels of protection strength. Here, $\eta$ above generated and protected images indicates the budget, i.e., the protection strength.
}
\vspace{-2mm}
\label{fig:perturbation_strength}
\end{figure}

\begin{figure}[t]
\centering
\newcommand{\w}{46mm}
\newcommand{\ww}{25mm}
\newcommand{\www}{5mm}
\newcommand{\h}{31.5mm}
\newcommand{\hh}{18mm}
\centering
\includegraphics[height=\h]{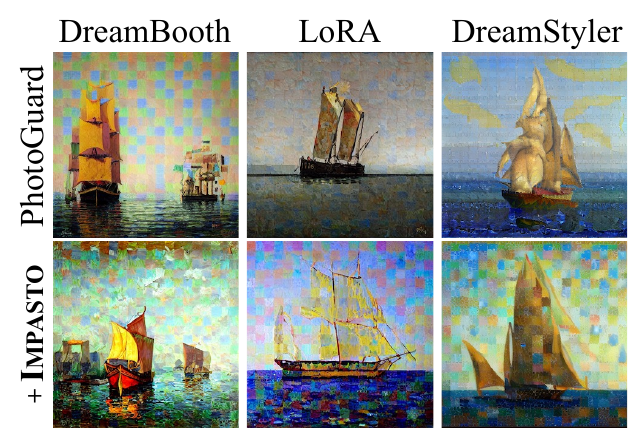}
\includegraphics[height=\h]{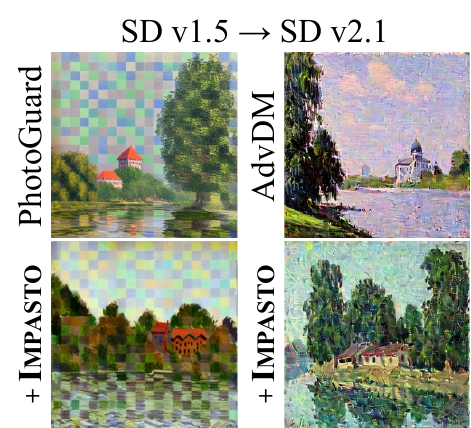}\hfill
\\
\makebox[\w][c]{(a) Personalization methods}
\makebox[\www][c]{}
\makebox[\ww][c]{(b) Diffusion models}\hfill
\\
\vspace{1mm}
\includegraphics[height=\hh]{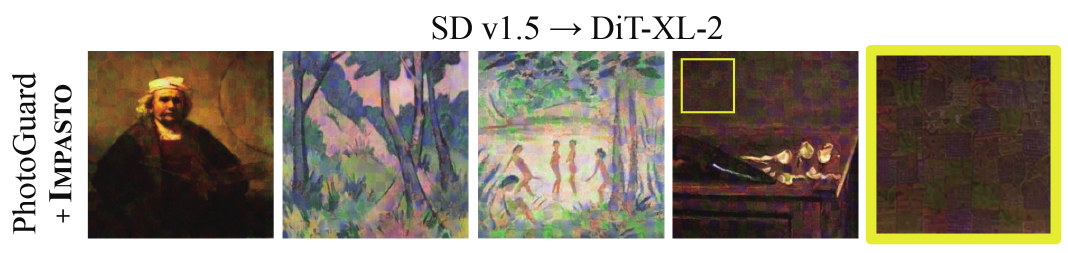}\hfill
\\
\makebox[\w][c]{(c) Diffusion models with transformers}
\caption{
\textbf{Generalization.} Generated images when various diffusion models fine-tuned with the protected images.
Note: For (a) and (b), protected image in Fig.~\ref{fig:qual_comp} (left) is used as the input for fine-tuning.
\ours\ does not impede baselines' generalization abilities on \textbf{(a)} different personalization methods; DreamBooth, LoRA, and DreamStyler, \textbf{(b)} model black-box scenario for diffusion models; SD v1.5 $\rightarrow$ v2.1, and \textbf{(c)} model black-box scenario for DiT; SD v1.5 $\rightarrow$ DiT-XL-2. }
\vspace{-2mm}
\label{fig:generalization}
\end{figure}

To demonstrate this, we match the protected image quality of PhotoGuard to PhotoGuard+\ours\ and compare the protection performance in both clean (no countermeasure) and blur scenarios (\Tref{tab:suppl_countermeasure}). In this case, which with similar imperceptibility, \ours\ shows better performance under blur, implying that such degradation is an inevitable limitation of using weaker perturbations rather than a flaw in our model design. Note that the performance gap narrows under blur because \ours\ applies very weak perturbations to more noticeable regions, potentially causing minor vulnerabilities. Nevertheless, in all scenarios—whether clean or under countermeasures—\ours\ demonstrates an improved trade-off between protected image quality and protection performance, highlighting its practical effectiveness.

\begin{table}[t]
\centering
\caption{\textbf{Comparison on SD with LoRA weights} of PhotoGuard with \ours\ at a similar level of protected image quality.}
\footnotesize
\setlength\tabcolsep{11pt}
\begin{tabular}{cccc}
\hline
\multirow{2}{*}{Method} & \multicolumn{3}{c}{Protection Performance} \\
\cline{2-4}

& NIQE ($\uparrow$) & BRISQUE ($\uparrow$) & FID ($\uparrow$) \\
\hline\hline
Pretrained SD & 5.620 & 11.46 & 157.7 \\
\hline
SD w/ LoRA$^{1}$  & 5.615 & 11.47 & 156.8 \\
SD w/ LoRA$^{2}$  & 5.623 & 11.38 & 155.5 \\
\hline
\end{tabular}
\begin{tablenotes}
\scriptsize	
\item LoRA$^{1}$: https://civitai.com/models/142551/fashion-manga-style-lora
\item LoRA$^{2}$: https://civitai.com/models/7801/oil-painting-style-lora 
\end{tablenotes}
\label{table:comp_sd_lora}
\end{table}

\begin{figure*}[t]
\centering
\begin{subfigure}{.45\textwidth}
    \includegraphics[width=\linewidth]{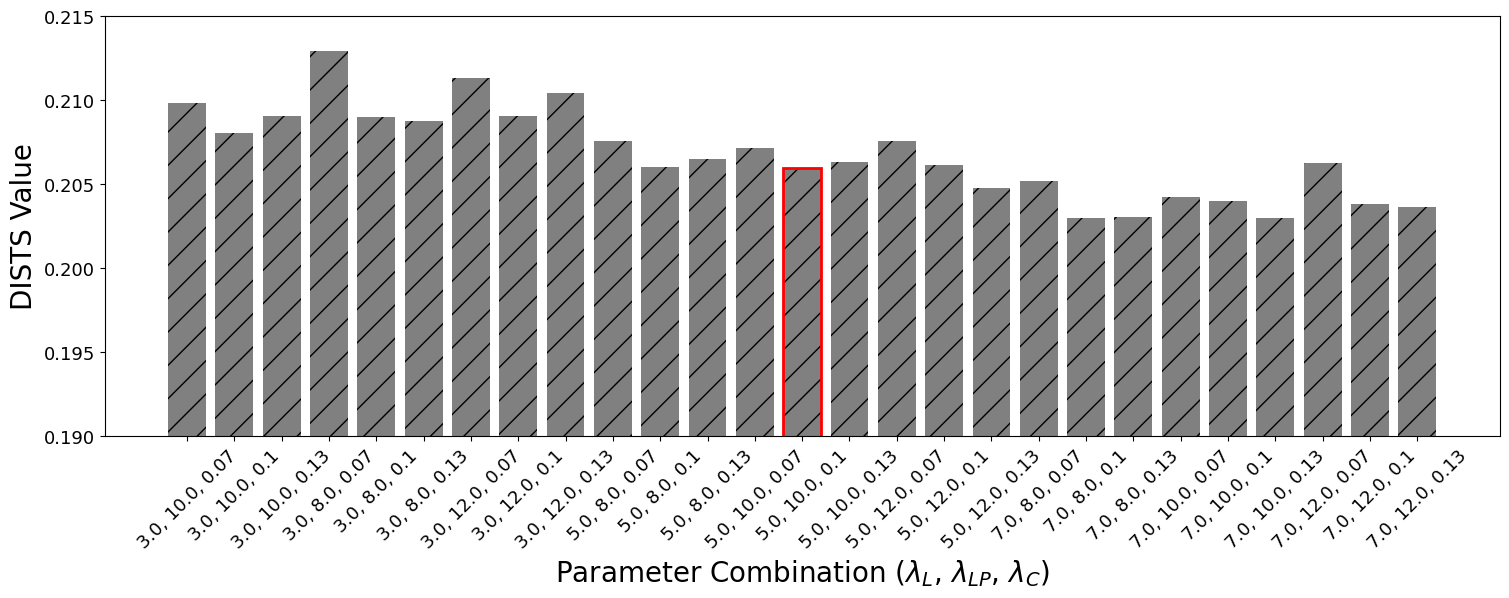}
    \caption{Results of DISTS}
\end{subfigure}
\quad
\begin{subfigure}{.45\textwidth}
    \includegraphics[width=\linewidth]{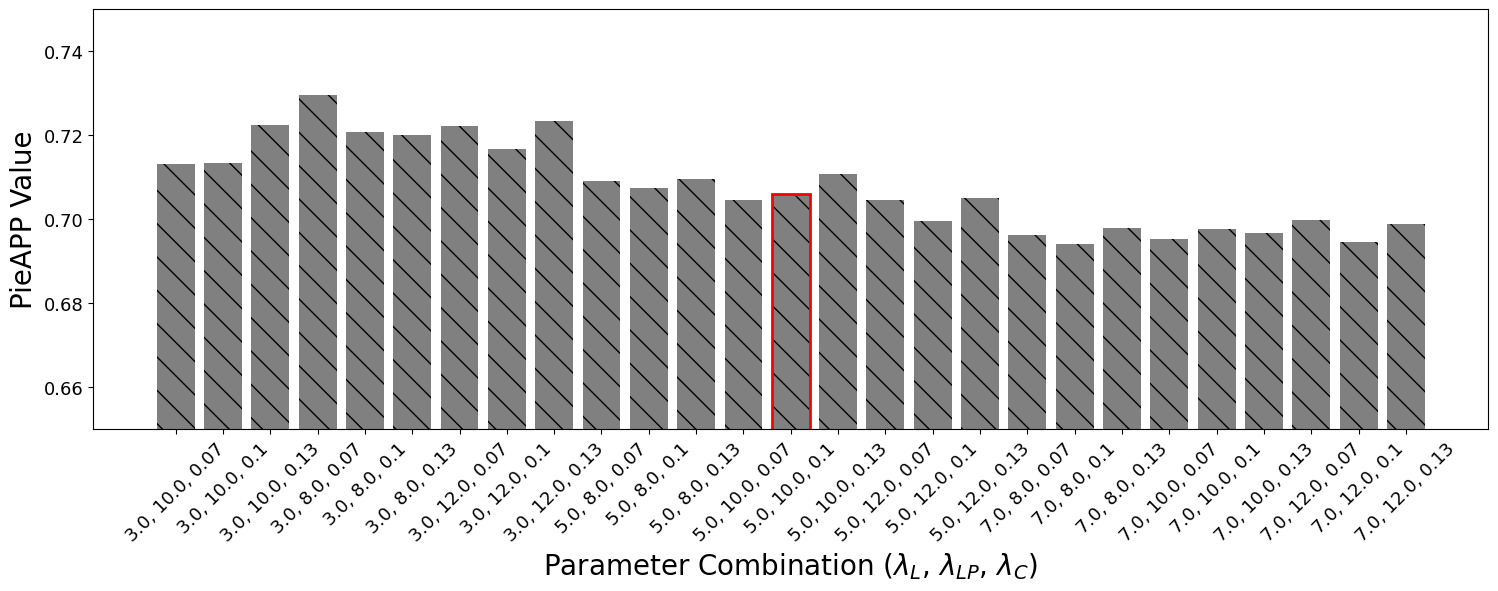}
    \caption{Results of PieAPP}
\end{subfigure}
\\
\vspace{1mm}
\begin{subfigure}{.45\textwidth}
    \includegraphics[width=\linewidth]{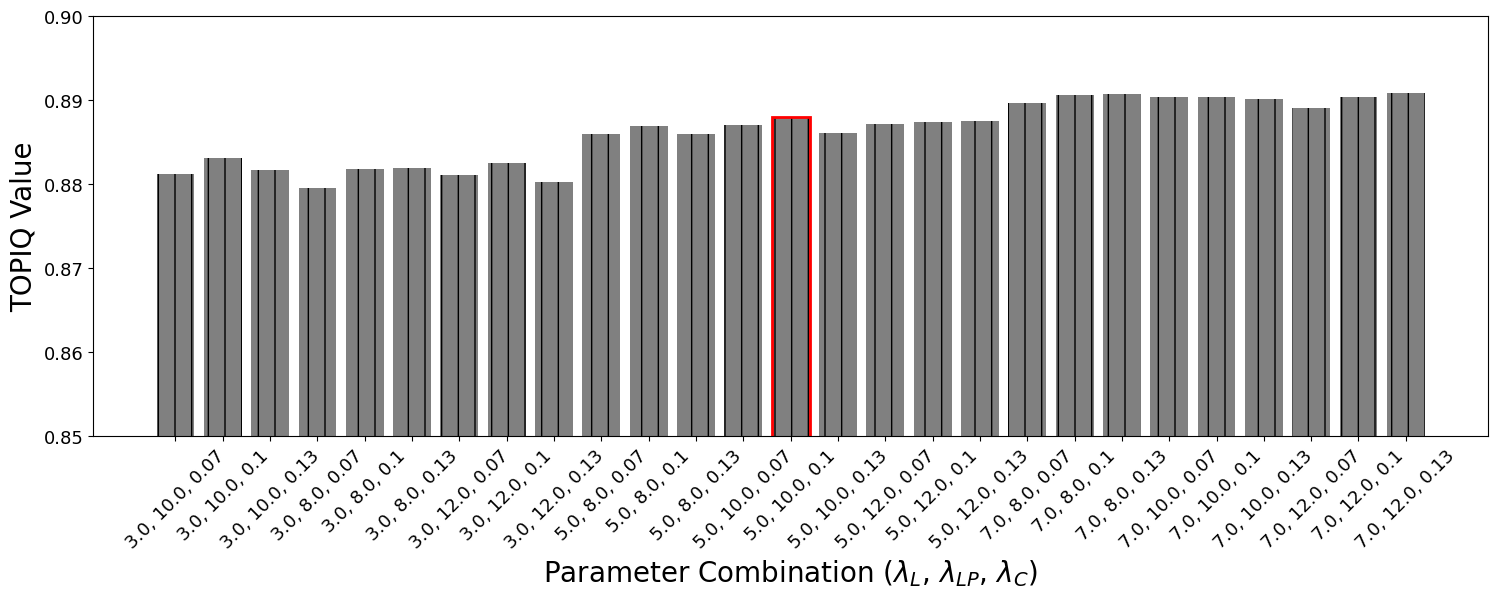}
    \caption{Results of TOPIQ}
\end{subfigure}
\quad
\begin{subfigure}{.45\textwidth}
    \includegraphics[width=\linewidth]{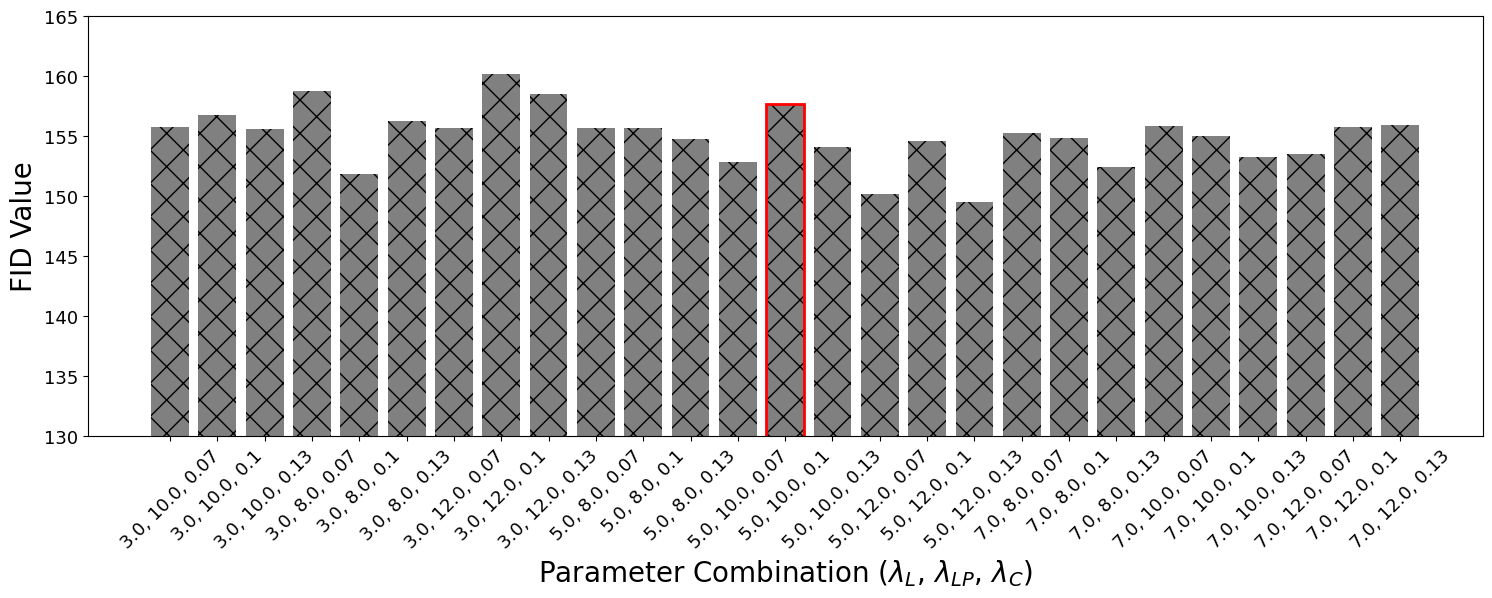}
    \caption{Results of FID}
\end{subfigure}

\caption{\textbf{Hyperparameters analysis.} For PhotoGuard with \ours, the combinations of three types of loss weights were evaluated for protection performance using FID and imperceptibility performance via DISTS, PieAPP, and TOPIQ metrics. The bar with the red border represents the results of the base settings.}
\label{fig:hyperparameters_analysis}
\end{figure*}

\smallskip
\noindent\textbf{Degradation tendency.}
Our work aims to enhance invisibility while preserving protection performance compared to baselines.
To minimize perceptible degradation, the perturbation signal must be both effective and minimally intrusive.
Thus, unlike adversarial attacks not tailored for invisibility, \ours\ does not induce severe visual distortions.
However, increasing the perturbation strength results in pronounced structural changes, as shown in Figure~\ref{fig:perturbation_strength}.
Notably, both PhotoGuard and AdvDM with \ours\ exhibit stronger reproduction of their characteristic patterns when the perturbation strength is increased (see yellow box regions).

\smallskip
\noindent\textbf{Generalization.}
We broaden the evaluation when applied to other personalization methods.
We adopt LoRA~\cite{hu2021lora}, a standard personalization method, and DreamStyler~\cite{ahn2023dreamstyler}, known for its effectiveness in style adaptation.
As shown in \Fref{fig:generalization}a, \ours\ maintains robust protection effectiveness.
In style protection, generalization robustness against unknown models is also a crucial aspect.
To examine this, we compare by optimizing on SD v1.5 and testing on SD v2.1.
As illustrated in \Fref{fig:generalization}b, \ours\ successfully preserves the robustness of PhotoGuard and AdvDM.
Furthermore, we conduct tests on models incorporating the recently proposed diffusion transformers (DiT) \cite{peebles2023scalable}, and we observed that PhotoGuard (encoder-based approach) with \ours~effectively reproduces the intended target image, specifically the pattern of the grid (see \Fref{fig:generalization}c).
Overall, in all three facets of the black-box scenario; unknown personalization methods, diffusion models, and diffusion models with transformers, \ours\ not only maintains the generalization ability but also effectively improves protection images' quality.

In addition, we conducted supplementary experiments to verify the preservation of protection performance in configurations combining pretrained SD v1.5 with finetuned weights.
In this experiments, we utilize LoRA models$^{1,2}$ that had been appropriately finetuned on other datasets.
We evaluate whether protected images optimized for the pretrained SD maintain style protection in the finetuned versions (SD with LoRA)
As listed in Table~\ref{table:comp_sd_lora}, it can be observed that the protection performance in terms of NIQE, BRISQUE, and FID is maintained even in the LoRA-integrated configuration.

\smallskip
\noindent\textbf{Hyperparameters analysis.}
We investigate the impact of the loss weights assigned to each component of the loss function.
As described in Sec.\ref{sec:suppl_implementation_details}, we use the following hyperparameters in Eq.\ref{eq:impaso}: $\lambda_L = 5.0$, $\lambda_{LP} = 10.0$, and $\lambda_C = 0.1$.
We evaluate imperceptibility and protection performance across 27 combinations by varying the three parameters at three levels each (see Figure~\ref{fig:hyperparameters_analysis}).
In terms of imperceptibility metrics (i.e., DISTS, PieAPP, and TOPIQ), increasing $\lambda_L$ generally improves performance.
For FID scores, we observe noticeable fluctuations, but the default settings achieve a good balance between imperceptibility and protection.
Based on this experiment, we selected the optimal hyperparameters via grid search on the validation set.

Additionally, we conduct a sensitivity analysis to evaluate the effect of diffusion process parameters (i.e., $t$) on the DAP approximation used in IMPASTO.
In our default design, we fix the timestep at $t=5$ out of $T=25$ to capture local perturbation effects.
This partial denoising balances computational efficiency with realism by approximating how models like DreamBooth use early-stage latent representations during fine-tuning.
To assess robustness, we vary the timestep $t$ across several values (e.g., $t=4, 5, 6$) and observe that the imperceptibility and protection performance remained consistent.

\begin{figure*}[t]
\centering
\includegraphics[width=\linewidth]{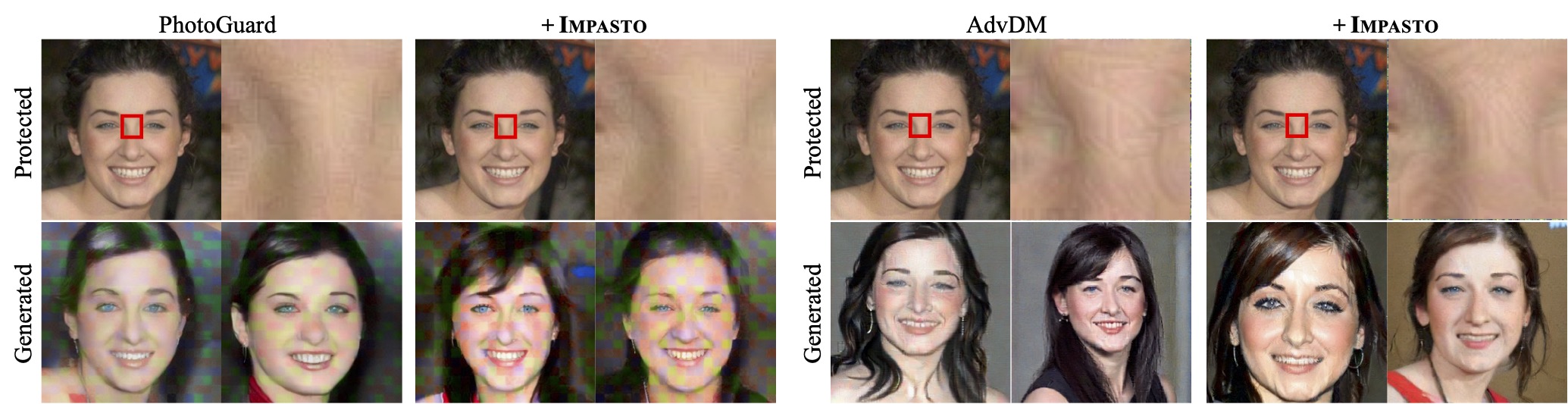}
\caption{
\textbf{Qualitative comparison} on facial dataset.
}
\label{fig:qual_comp_face}
\vspace{-2mm}
\end{figure*}

\smallskip
\noindent\textbf{\ours\ in other domain.}
Although \ours\ is initially proposed to prevent style imitation, its applicability can be extended beyond other domains or applications.
To validate this, we conduct protection on two facial datasets, CelebA-HQ~\cite{karras2017progressive} and VGGFace2~\cite{cao2018vggface2}.
As demonstrated in \Tref{table:comp_fix_protection_face} and \Fref{fig:qual_comp_face}, adopting \ours\ in these natural domains can also enhance the quality of protected images without compromising the protection performance of baseline models.
In addition to facial datasets, we also conducted robustness evaluations on iCartoonFace~\cite{zheng2020cartoon}, which contains a wide range of cartoon-style images, and the Stanford Online Products (SOP) dataset~\cite{song2016deep}, which consists of real-world object images.
As listed in the \Tref{table:comp_new_data}, the proposed IMPASTO demonstrates acceptable performance in terms of both DISTS and FID on diverse types of cartoon and real-world object images.
This broad applicability highlights \ours's versatility and potential as a universal tool for protecting users' copyright and preventing serious threats of deepfakes.

\begin{table*}[t]
\centering
\caption{\textbf{Quantitative comparison} on facial datasets.}
\small
\setlength\tabcolsep{10pt}
\begin{tabular}{cccccccc}
\hline
\multirow{2}{*}{Dataset} & \multirow{2}{*}{Method} & \multicolumn{3}{c}{Protected Image Quality} & \multicolumn{3}{c}{Protection Performance} \\
\cline{3-8}
& & DISTS ($\downarrow$) & PieAPP ($\downarrow$) & TOPIQ ($\uparrow$) & NIQE ($\uparrow$) & BRISQUE ($\uparrow$) & FID ($\uparrow$) \\
\hline
\hline
\multirow{4}{*}{CelebA-HQ} & PhotoGuard   & 0.280 \equal{0.000} & 0.379 \equal{0.000} & 0.878 \equal{0.000} & 5.031 & 15.36 & 233.1 \\
& + \ours                                       & 0.266 \good{0.014}  & 0.376 \good{0.003}  & 0.880 \good{0.002}  & 5.161 & 14.13 & 234.6 \\
\cline{2-8}
& AdvDM                            & 0.213 \equal{0.000} & 0.573 \equal{0.000} &  0.832\equal{0.000} & 4.051 & 10.05 & 278.8 \\
& + \ours                          & 0.195 \good{0.018}  & 0.522 \good{0.051}  &  0.854\good{0.022}  & 4.080 & 10.59 & 273.6 \\
\hline
\multirow{4}{*}{VGGFace2} & PhotoGuard & 0.270 \equal{0.000} & 0.413 \equal{0.000} & 0.870 \equal{0.000} & 5.916 & 19.73 & 253.9 \\
& + \ours                                    & 0.255 \good{0.015}  & 0.413 \equal{0.000} & 0.868 \bad{0.002}  & 5.746 & 19.39 & 257.0 \\
\cline{2-8}
& AdvDM                            & 0.206 \equal{0.000} & 0.589 \equal{0.000} & 0.821 \equal{0.000} & 3.977 & 8.43 & 292.9 \\
& + \ours                          & 0.187 \good{0.019}  & 0.543 \good{0.043}  & 0.846 \good{0.025} & 3.986 & 9.86 & 289.8 \\
\hline
\end{tabular}
\label{table:comp_fix_protection_face}
\end{table*}

\begin{table}[t]
\centering
\caption{
\textbf{Quantitative comparison} on cartoon and object datasets.}
\scriptsize
\setlength\tabcolsep{10pt}
\begin{tabular}{cccc}
\hline
\multirow{2}{*}{Dataset} & \multirow{2}{*}{Method} & \multicolumn{1}{c}{Image Quality} & \multicolumn{1}{c}{Prot. performance} \\
\cline{3-4}
& & DISTS ($\downarrow$) & FID ($\uparrow$) \\
\hline
\hline
\multirow{4}{*}{iCartoonFace} & PhotoGuard   & 0.242 \equal{0.000} & 159.4 \\
& + \ours                                    &  0.205 \good{0.037} & 158.2 \\
\cline{2-4}
& AdvDM                            & 0.234 \equal{0.000}  &  156.3 \\
& + \ours                          & 0.211 \good{0.023} &  157.9 \\
\hline
\multirow{4}{*}{SOP dataset} & PhotoGuard & 0.274 \equal{0.000} & 248.3 \\
& + \ours                                    & 0.258 \good{0.016} & 250.2 \\
\cline{2-4}
& AdvDM                            & 0.215 \equal{0.000} & 281.5 \\
& + \ours                          &  0.199 \good{0.016} & 277.2 \\
\hline
\end{tabular}
\label{table:comp_new_data}
\end{table}
\section{Conclusion}
We propose \ours\ to prevent style imitation with a perceptual-oriented approach. By incorporating perceptual and difficulty-aware protections, along with a perceptual constraint bank, \ours\ significantly enhances the quality of protected images within existing protection frameworks. Through a series of benchmarks, we demonstrate that applying \ours\ to baseline protection methods improves imperceptibility without compromising protection performance or robustness.

\smallskip
\noindent\textbf{Limitations.} 
All the current protections mostly adopt adversarial perturbations, which can hamper usability due to the extensive time required for the optimization.
Even accessible software~\cite{shan2023glaze} takes 30-60 mins to protect a $512^2$ image on CPU.
Addressing this time constraint is a challenge that future research should aim to overcome.

\bibliographystyle{IEEEtran}
\bibliography{main}
 
\begin{IEEEbiography}[{\includegraphics[width=1in,height=1.25in]{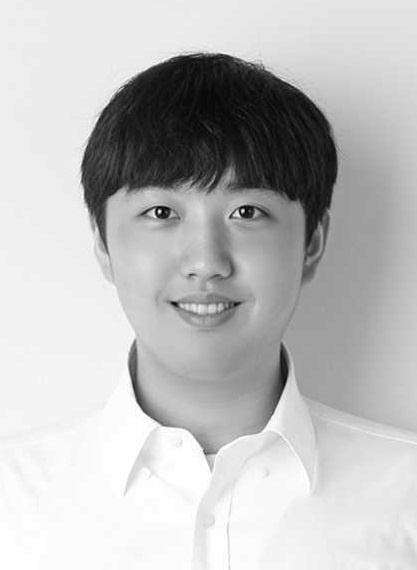}}]{\textbf{Namhyuk Ahn}} (Member, IEEE) received the B.S. degree in media and the Ph.D. degree in artificial intelligence from Ajou University, in February 2016 and August 2021, respectively. He was an AI Researcher with Naver Webtoon AI, from 2021 to 2024. Since September 2024, he has been an Assistant Professor with the Department of Electrical and Electronic Engineering, Inha University. His research interests include generative AI and computer vision, with a focus on applications that leverage multimodal models, such as text-to-image and language models.
\end{IEEEbiography}

\begin{IEEEbiography}[{\includegraphics[width=1in,height=1.25in]{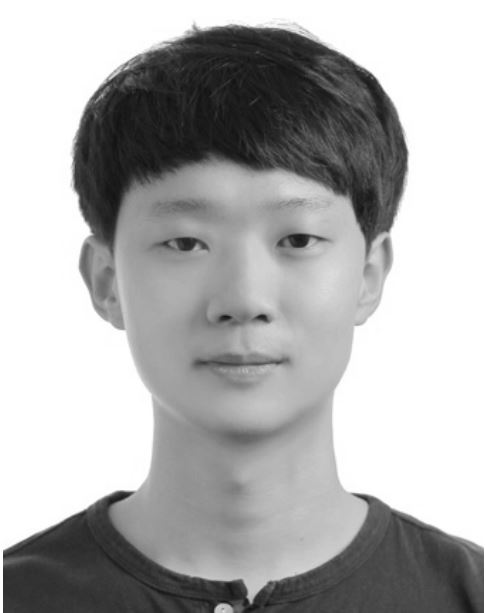}}]{\textbf{Wonhyuk Ahn}} is an AI researcher at NAVER WEBTOON AI, South Korea. He earned his Ph.D. in School of Computing from the Korea Advanced Institute of Science and Technology (KAIST), where he was supervised by Professor Lee Heung‑kyu, and holds a B.S. from Ajou University. His research lies at the intersection of deep learning and multimedia content protection.
\end{IEEEbiography}

\begin{IEEEbiography}[{\includegraphics[width=1in,height=1.25in]{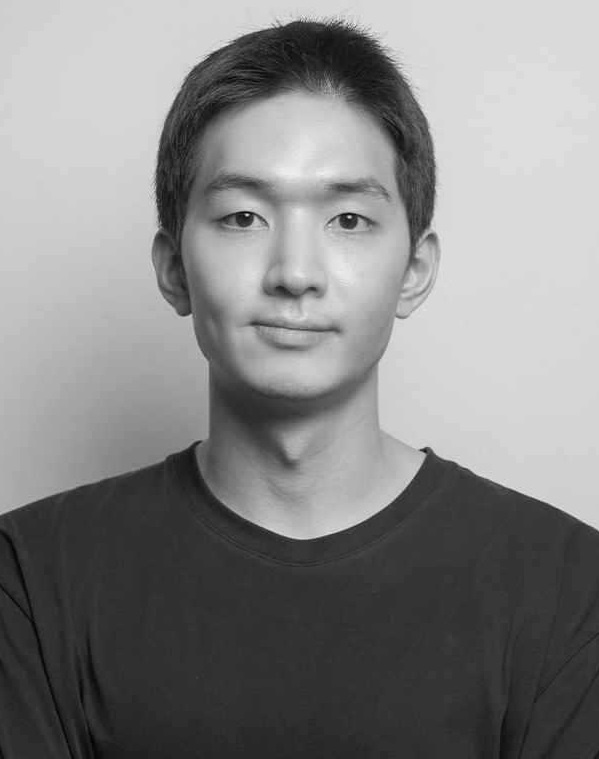}}]{\textbf{KiYoon Yoo}} received his Ph.D. from Seoul National University under Nojun Kwak’s guidance with research emphasizing the safety and robustness of language models, including areas such as adversarial attack and defense and watermarking. Currently, he builds on-device agents for gameplay, enabling real-time, context-aware interactions and strategic cooperation with players.
\end{IEEEbiography}

\begin{IEEEbiography}[{\includegraphics[width=1in,height=1.25in]{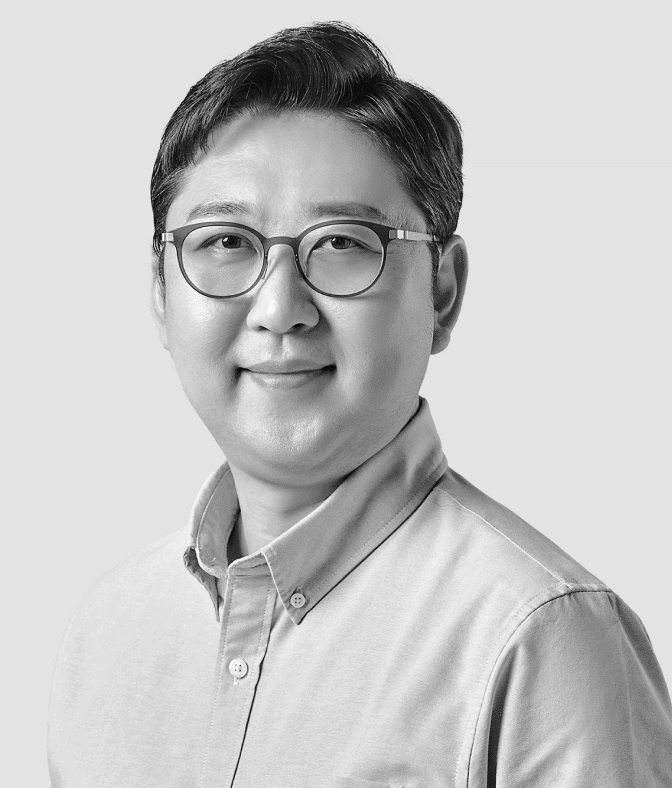}}]{\textbf{Daesik Kim}} is the Head of AI\&Data for WEBTOON Entertainment and its subsidiaries. He began to oversee NAVER WEBTOON’s artificial intelligence organizations in November 2020, in a role that later expanded to oversee both data and AI teams in November 2022. He was appointed Vice President of NAVER WEBTOON in 2022.
Before joining NAVER WEBTOON, He served as CEO of AI startup V.DO Inc. from September 2017 to December 2019, where he also oversaw business management and deep learning computer vision project management as CTO. He received his B.S. degree in Industrial and Management Engineering from POSTECH and his Ph.D. in Intelligence and Information (machine learning and deep learning research) from Seoul National University.
\end{IEEEbiography}

\begin{IEEEbiography}[{\includegraphics[width=1in,height=1.25in]{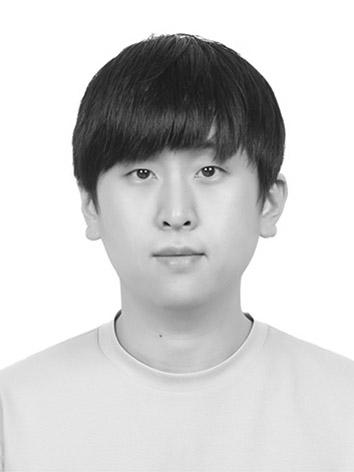}}]{\textbf{Seung-Hun Nam}} received the B.S. degree in Information and Communication Engineering from Dongguk University in 2013, and the M.S. and Ph.D. degrees in Computer Science from the Korea Advanced Institute of Science and Technology (KAIST) in 2015 and 2020, respectively.
He is currently working as an AI researcher at NAVER WEBTOON AI.
His research interests include various aspects of digital watermarking, multimedia forensics, and multimedia content protection.

\end{IEEEbiography}

\vfill

\appendix

\section{Method Details}
\subsection{JND Estimations}
In our study, we investigated five widely-used JND models, briefly illustrated below.

\begin{itemize}
    \item \textbf{Luminance adaptation (LA)}: Perturbations are less visible in regions of very low or high luminance and more noticeable in moderate lighting conditions~\cite {jarsky2011synaptic}. Hence, we modulate protection strength based on pixel luminance with a fixed adaptation model.
    
    \item \textbf{Contrast masking (CM):} Perturbations can seamlessly blend into regions with intricate textures, while they leave distinct traces on flat surfaces. To simulate this, we utilize the luminance contrast (or change) of a region to measure the complexity of the pixel~\cite{legge1980contrast,wu2013pattern}.

    \item \textbf{Contrast sensitivity function (CSF):} Given the band-pass characteristics of the human visual system, we utilize a frequency-based JND model~\cite{wei2009spatio}. The human eye is receptive to signals at modulated frequencies while exhibiting insensitivity to high-frequency components. Consequently, perturbations overlaid on high-frequency signals (e.g. edges) are less perceptible.
    
    \item \textbf{Standard deviation:} To assess the spatial structure of an image, we calculate the standard deviation of local image blocks, inspired by SSIM~\cite{wang2004image}. This measures the image's structural complexity, which correlates with the sensitivity to subtle perturbations.
    
    \item \textbf{Entropy:} The entropy of an image block is computed to quantify the amount of information or complexity within a local region~\cite{wu2013just}.
\end{itemize}

\noindent We also provide a more comprehensive elucidation of the fundamental rationale of these estimations and delve into the specifics of their computations:

\smallskip
\noindent\textbf{Luminance adaptation.}
The visibility thresholds within the human visual system vary to luminance levels~\cite{jarsky2011synaptic}.
For instance, our eye is more sensitive under moderate lighting conditions, while discrimination is challenging in a very dark light.
Similarly, we observed that perturbations are less noticeable in regions with extremely low or high luminance, prompting an increase in protection strength within these areas.
To implement this concept, luminance adaptation is computed following Chou and Li~\cite{chou1995perceptually} as in the below equation.
\begin{equation}
\setlength{\medmuskip}{1.5mu}
LA(\mathbf{x}) = 
\begin{cases}
17 \times (1 - \sqrt{\frac{B(\mathbf{x})}{127}}) + 3, & \text{if } B(\mathbf{x}) \leq 127 \\
\frac{3}{128} \times (B(\mathbf{x})-127) + 3, & \text{otherwise},
\end{cases}
\end{equation}
where $B(\mathbf{x})$ is the background luminance, which is calculated as the mean luminance of a local $3\times3$ block.
With this formulation, the sensitivity (the inverse of LA) peaks at luminance ranging from 100 to 200.
As luminance approaches an extremely low value (darker area), human perception fails to discern minor variations.
Conversely, at high luminance levels, the detection ability becomes progressively more difficult, albeit less so compared to the darker regions.

\smallskip
\noindent\textbf{Contrast masking.}
It is obvious that stimuli become less perceptible against patterned, non-uniform backgrounds.
In our context, perturbations can seamlessly blend into regions with intricate textures, while they leave distinct traces on flat surfaces.
To simulate this effect, we first compute the luminance contrast~\cite{legge1980contrast} by convolving an input image with four directional filters.
This process highlights the complexity of a region by contrasting it with its immediate surroundings.
Subsequently, contrast masking is established by mapping the luminance contrast onto a logarithmic curve~\cite{wu2013pattern}.
\begin{equation}
CM(\mathbf{x}) = 0.115 \times \frac{16\times LC(\mathbf{x})^{2.4}}{LC(\mathbf{x})^2+26^2}, 
\end{equation}
where luminance contrast $LC(\mathbf{x})$ is obtained via four directional filters as $LC(\mathbf{x}) = \frac{1}{16}\times \text{max}_{k=1,\dots, 4}|\mathbf{x} * \nabla_k|$ following Wu et al.~\cite{wu2019survey}.

\smallskip
\noindent\textbf{Contrast sensitivity function.}
The human visual system exhibits a band-pass response to spatial frequency and CSF represents a function of how our eye is sensitive to the contrast of signals at various spatial frequencies.
High spatial frequencies correspond to rapid changes in image details, and contrast sensitivity is optimal at moderate spatial frequencies.
Beyond a certain frequency threshold, known as the resolution limit, our eyes are unable to detect changes.
Upon this concept, several studies have introduced and modified the CSF~\cite{larson2010most,mitsa1993evaluation,damera2000image,mannos1974effects}; the CSF model $H(f, \theta)$ is given below.
\begin{equation}
H =
\begin{cases}
2.6 \times (\alpha+\beta f_\theta) e^{-(\beta f_\theta)^{1.1}}, & \text{if } f \geq 7.8909 \\
0.981, & \text{otherwise,}
\end{cases}
\end{equation}
where $\alpha=0.0192, \beta=0.114$ and $f$ is the radial spatial frequency, measured in cycles per degree of visual angle (c/deg).
The variable $\theta$, ranging from $[-\pi, \pi]$, denotes the orientation.
Additionally, $f_\theta = f / [0.15 \times \text{cos}(4\theta) + 0.85]$ accounts for the oblique effect~\cite{larson2010most}.

\noindent The CSF model is now applied to an image in the frequency domain as $\mathbf{x}_{csf} = \mathcal{F}^{-1} [ H(u, v) \times \mathcal{F}(\check{\mathbf{x}})]$, where $\mathcal{F}[\cdot]$ and $\mathcal{F}^{-1}[\cdot]$ represent the DFT and its inverse, respectively.
Here, $H(u, v)$ is the DFT-version of $H(f, \theta)$, with $u, v$ being the DFT indices.
The transformation from $H(f, \theta)$ to $H(u, v)$ is elaborated in Larson et al.~\cite{larson2010most}.
Prior to applying the CSF model, an input image $\mathbf{x}$ is converted into a perception-adjusted luminance form to reflect the non-linear relationship between digital pixel values and physical luminance~\cite{larson2010most}.
This is achieved by calibrating $\mathbf{x}$ to the settings of an sRGB display and converting it into perceived luminances, which indicates the relative lightness: $\check{\mathbf{x}} = \sqrt[3]{(0.02874\mathbf{x})^{2.2}}$.

\smallskip
\noindent\textbf{Standard deviation.}
It has served as a crucial metric for quantifying the structural information of an image, both in pixel space~\cite{wang2004image} and feature space~\cite{ding2020image}.
Given that perturbations tend to be less noticeable in areas exhibiting high levels of change, (as discussed in contrast masking), we focus on these sudden changes, interpreting them as structural components.
For this purpose, we compute the block-wise standard deviation in pixel space using a $9\times9$ local block.

\smallskip
\noindent\textbf{Entropy.}
Similar to standard deviation, block-wise entropy is calculated using a $9\times9$ local window.
This approach is based on the concept that the complexity of a local region influences the detectability of subtle perturbations.
The higher the local entropy, the more intricate the region, potentially rendering minor perturbations less perceptible.

\smallskip
\noindent\textbf{Post-processing}
All the JND estimations are min-max normalized and then inverted by subtracting from one.
When constructing a perceptual map $\mathcal{M}$ with JND estimations, we encountered issues with some JNDs displaying extremely skewed distributions.
Additionally, the distributions of JNDs often do not align with each other, posing challenges in their combination.
While standardization aligns the distributions, we found that the continuous JND values provide weak signals as masks (e.g. on average, the protection strength drops to 50\% compared to the original).
To give a more distinct signal depending on the JND values, we discretize the scores by quantizing the JND values.
This involves calculating JND quantiles, where the first quantile is set to 1.0.
For each subsequent quantile, we multiply by a factor of $\beta$ to decrease the value in a discrete manner.
We use $\beta = 0.85$, resulting in quantized JND values across four quantiles as [1.0, 0.85, 0.7225, 0.6141]. The value of $\beta$ was selected via grid search on the validation set to achieve the best trade-off between protection performance and imperceptibility.
This allows for a more nuanced adjustment of the protection intensity while accommodating the varied distributions of JND estimations.

\subsection{Perceptual Constraints}
\label{sec:suppl_constraints}

\noindent\textbf{Masked LPIPS.} 
We use AlexNet~\cite{krizhevsky2012imagenet} as the backbone network for the LPIPS loss, adhering to the parameter settings in the official LPIPS implementation.

\smallskip
\noindent\textbf{Masked low-pass.}
In this constraint, we utilize $\text{LP}(\cdot)$, a reconstruction function focusing on the low-frequency component.
Inspired by Luo et al.~\cite{luo2022frequency}, we implement a DWT-based reconstruction module.
Given an input image $\mathbf{x}$, DWT decomposes it into one low-frequency component and three high-frequency components, as expressed by the following equation.
\begin{equation}
\begin{aligned}
\mathbf{x}_{ll} &= \mathbf{LxL}^T,\;\; \mathbf{x}_{lh} = \mathbf{HxL}^T, \\
\mathbf{x}_{hl} &= \mathbf{LxH}^T,\;\; \mathbf{x}_{hh} = \mathbf{HxH}^T.
\end{aligned}
\end{equation}
$\mathbf{L}$ and $\mathbf{H}$ represent the low-pass and high-pass filters of an orthogonal wavelet, respectively.
To reconstruct an image using only its low-frequency component, we input $\mathbf{x}_{ll}$ alone into the inverse DWT function.
Thus, $\text{LP}(\mathbf{x})$ is defined as:
\begin{equation}
\text{LP}(\mathbf{x}) \;=\; \mathbf{L}^T\mathbf{x}_{ll}\mathbf{L} \;=\; \mathbf{L}^T(\mathbf{LxL}^T)\mathbf{L}.
\end{equation}

\smallskip
\noindent\textbf{CLIP.} 
As outlined in Eq. 12, \ours\ focuses on maximizing the feature distance between the protected image and the prompt $C=$ ``Noise-free image''.
While it is conceivable to minimize the distance using a `bad' prompt (e.g. $C=$ ``noisy image'') or to integrate both `good' and `bad' prompts, we observed that all these show comparable performance. Therefore, we opted to employ the `good' prompt only in the CLIP-based constraint.



\section{Experimental Settings}
\noindent\textbf{Datasets.}
In our study, we chose the painting and cartoon domains, which are most significantly impacted by style imitation from diffusion models, and used them as benchmark datasets.
To create the ``painting'' dataset, we gather artworks from the WikiArt dataset~\cite{tan2018improved}.
This dataset includes selections from 15 artists, 10 artworks per each.
The ``cartoon'' dataset is sourced from the NAVER WEBTOON platform; images are cropped to contain only character depictions.
This comprises 15 cartoons with 10 cartoon character images each.
Following previous dataset preparation protocols~\cite{van2023anti}, each image was resized to 512$\times$512.
And note that as \ours\ does not require a separate training process, the data was used as benchmark set.

\begin{figure*}[t]
\centering
\includegraphics[width=\linewidth]{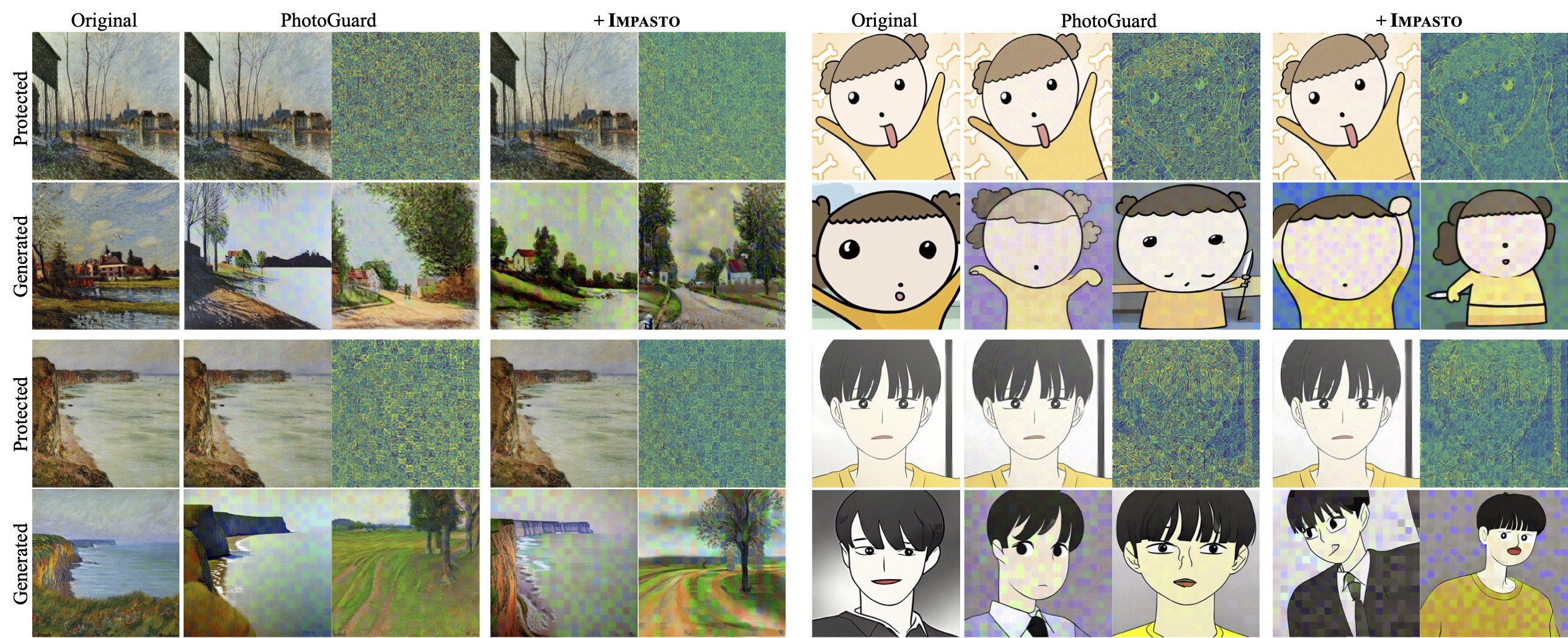}\\
\includegraphics[width=\linewidth]{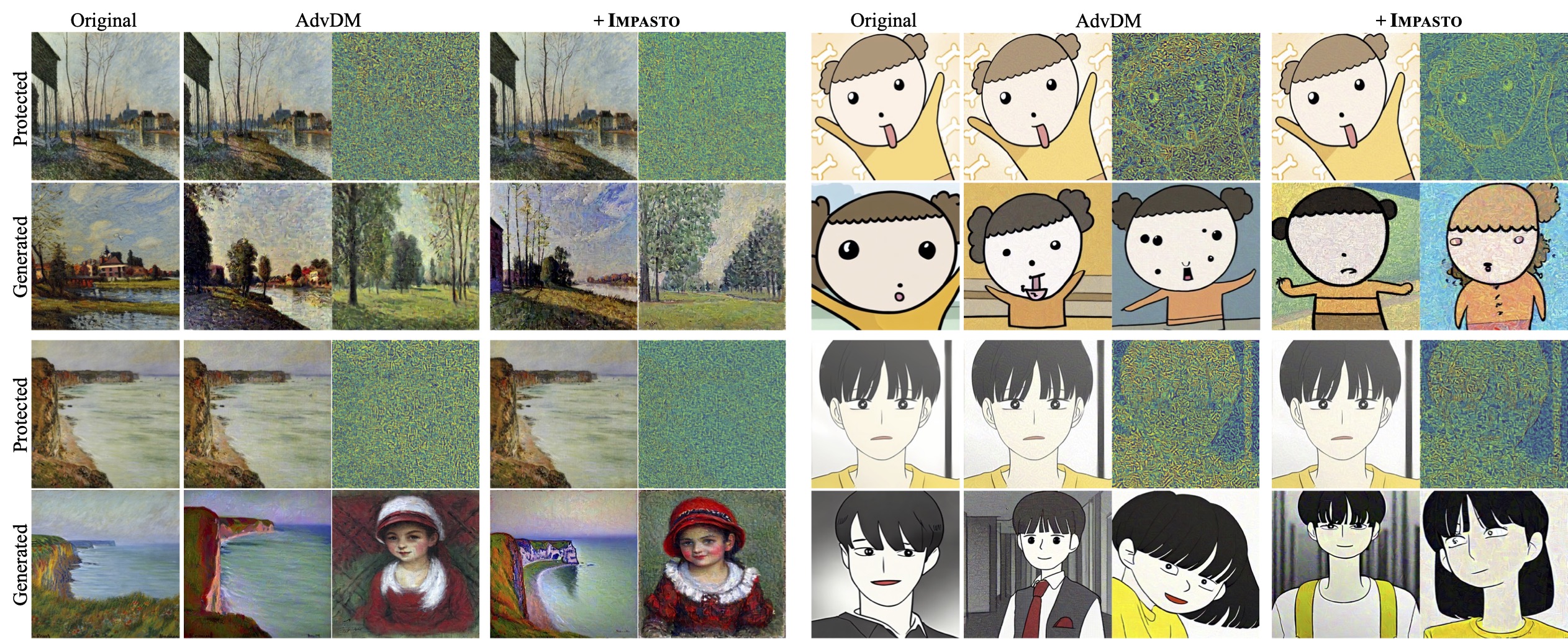}
\caption{
\textbf{Qualitative comparison.} We visualize the difference map ($\Delta$) between the protected and unprotected images. Yellowish color indicates different regions and bluish color show similar pixels. While maintaining comparable style protection efficacy (artifacts in the generated results), \ours\ significantly enhances the quality of the protected images.
}
\label{fig:qual_comp}
\end{figure*}

\begin{table*}[t]
\centering
\caption{\textbf{Quantitative comparison} of protection methods with and without \ours. Unlike Table I, we match the protected image quality to a similar level and compare the protection performance.}
\small
\setlength\tabcolsep{11pt}
\begin{tabular}{cccccccc}
\hline
\multirow{2}{*}{Dataset} & \multirow{2}{*}{Method} & \multicolumn{3}{c}{Protected Image Quality} & \multicolumn{3}{c}{Protection Performance} \\
\cline{3-8}
& & DISTS ($\downarrow$) & PieAPP ($\downarrow$) & TOPIQ ($\uparrow$) & NIQE ($\uparrow$) & BRISQUE ($\uparrow$) & FID ($\uparrow$) \\
\hline\hline
\multirow{4}{*}{Painting} & PhotoGuard   
& 0.181 & 0.364 & 0.896 & 4.306 \equal{0.000} & 20.99 \equal{0.00} & 277.6 \equal{0.0} \\
& + \ours                                
& 0.186 & 0.346 & 0.889 & 4.451 \good{0.145} & 22.02 \good{1.03} & 286.9 \good{9.3} \\
\cline{2-8}
& AdvDM                            
& 0.167 & 0.730 &  0.846 & 3.761 \equal{0.000} & 12.45 \equal{0.00} & 269.0 \equal{0.0} \\
& + \ours                          
& 0.164 & 0.503 & 0.842 & 3.947 \good{0.231} & 11.60 \bad{0.85} & 276.6 \good{7.6} \\
\hline
\multirow{4}{*}{Cartoon} & PhotoGuard 
& 0.249 & 0.782 & 0.797 & 5.037 \equal{0.000} & 10.19 \equal{0.00} & 155.9 \equal{0.0} \\
& + \ours                                   
& 0.241 & 0.894 & 0.779 & 5.480 \good{0.443} & 12.88 \good{2.69} & 162.5 \good{6.6} \\
\cline{2-8}
& AdvDM                            
& 0.241 & 0.776 & 0.775 & 4.802 \equal{0.000} & 10.95 \equal{0.00} & 153.5 \equal{0.0} \\
& + \ours                          
& 0.243 & 0.819 & 0.744 & 4.780 \bad{0.022} & 11.90 \good{0.95} & 155.9 \good{2.4} \\
\hline
\end{tabular}
\label{table:suppl_fix_quality}
\end{table*}

\section{Additional Results}

\noindent\textbf{Protection performance comparison.}
Here, unlike Table I, which compares the protected images’ quality under similar protection performance, we compare the protection performance while fixing the image quality. In \Tref{table:suppl_fix_quality}, \ours\ improves protection performance compared to the baseline in most metrics and scenarios. This was also shown in the quality-protection trade-off (Figure 7). These results indicate that attaching \ours\ to the protection baselines enables them to be better protection methods.

\begin{figure*}[htp]
\centering
\includegraphics[width=\linewidth]{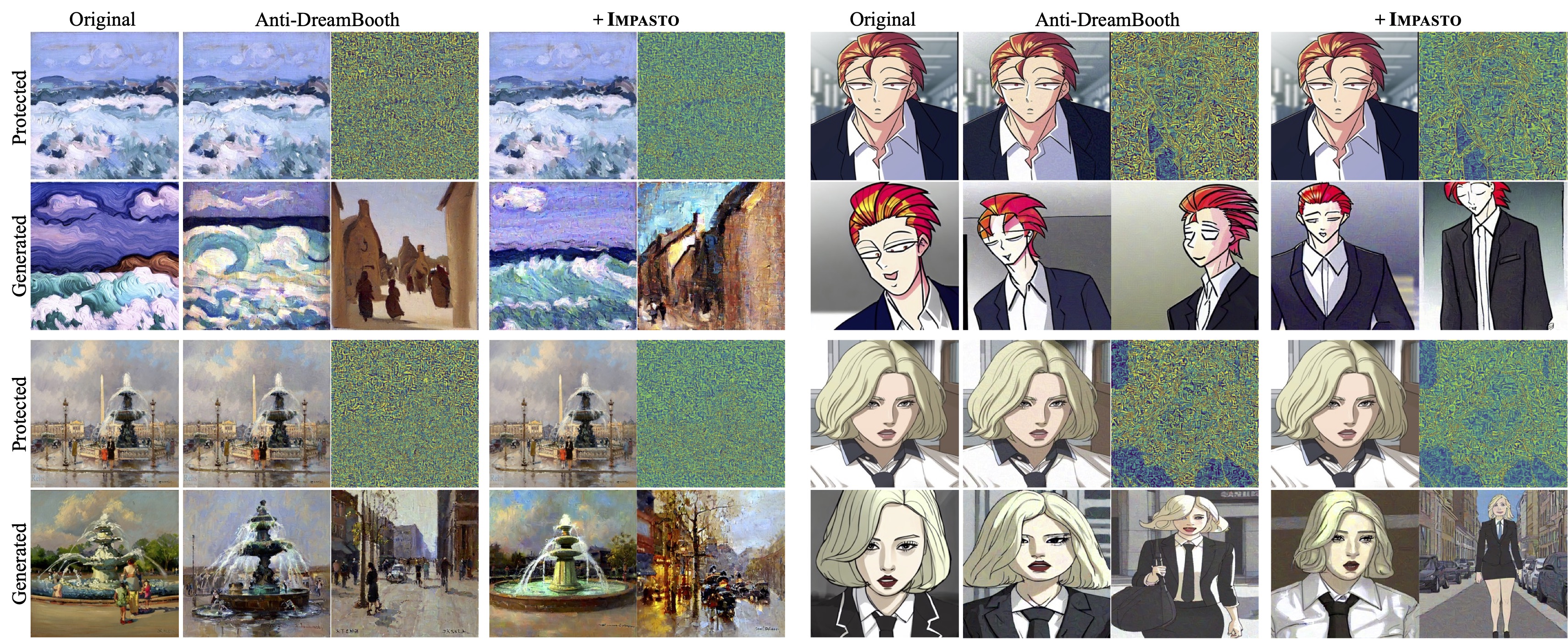}
\caption{\textbf{Qualitative comparison} of Anti-Dreambooth~\cite{van2023anti} with and without \ours. Best viewed in zoom.}
\label{fig:suppl_adb}
\end{figure*}

\begin{figure*}[htp]
\centering
\includegraphics[width=\linewidth]{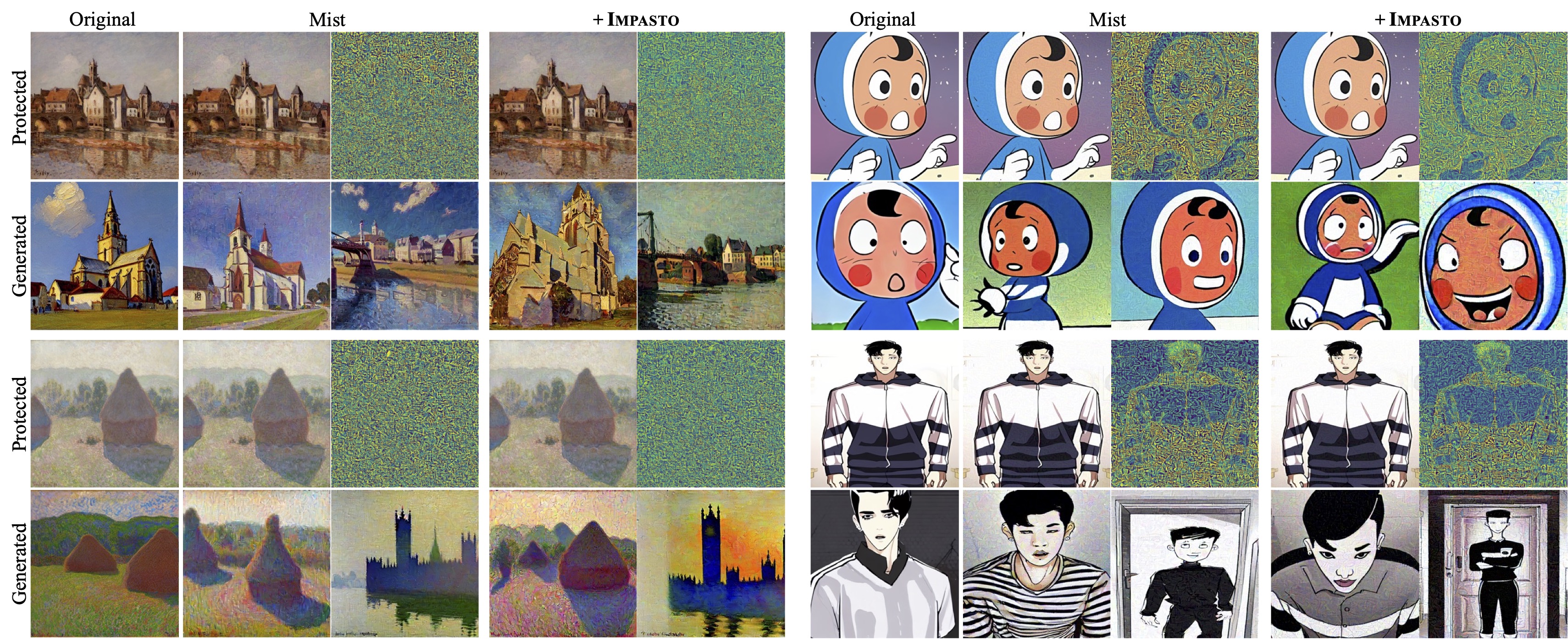}
\caption{\textbf{Qualitative comparison} of Mist~\cite{liang2023mist} with and without \ours. Best viewed in zoom.}
\label{fig:suppl_mist}
\end{figure*}

\smallskip
\noindent\textbf{Qualitative results.}
\Fref{fig:suppl_adb} and \ref{fig:suppl_mist} provide visual comparisons with and without \ours\ on Anti-DreamBooth~\cite{van2023anti} and Mist~\cite{liang2023mist}, respectively.


\end{document}